\documentclass[fleqn,10pt]{wlscirep}
\usepackage[utf8]{inputenc}
\usepackage[T1]{fontenc}
\usepackage{multirow}
\usepackage{multicol}
\usepackage{adjustbox}
\usepackage{hyperref}
\usepackage{url}
\usepackage{lscape}
\usepackage{fltpage}
\usepackage{xcolor} 
\usepackage{soul,color}
\usepackage{newclude}
\author[1,3,*]{Sharib {Ali}}
\author[1]{Mariia {Dmitrieva}}
\author[7]{Noha {Ghatwary}}
\author[8]{Sophia {Bano}}
\author[10]{Gorkem {Polat}}
\author[10]{Alptekin {Temizel}}
\author[11]{Adrian {Krenzer}}
\author[11]{Amar {Hekalo}}
\author[12]{Yun Bo {Guo}}
\author[12]{Bogdan {Matuszewski}}
\author[24]{Mourad {Gridach}}
\author[13]{Irina {Voiculescu}}
\author[14]{Vishnusai {Yoganand}}
\author[15]{Arnav {Chavan}}
\author[15]{Aryan {Raj}}
\author[16]{Nhan T. {Nguyen}}
\author[16]{Dat Q. {Tran}}
\author[17]{Le Duy {Huynh}}
\author[17]{Nicolas {Boutry}}
\author[18]{Shahadate Rezvy}
\author[19]{Haijian Chen}
\author[20]{Yoon Ho Choi}
\author[21]{Anand Subramanian}
\author[22]{Velmurugan Balasubramanian}
\author[18]{Xiaohong W. Gao}
\author[23]{Hongyu Hu}
\author[23]{Yusheng Liao}
\author[8]{Danail {Stoyanov}}
\author[9]{Christian Daul}
\author[5]{Stefano {Realdon}}
\author[6]{Renato {Cannizzaro}}
\author[4]{Dominique {Lamarque}}
\author[2]{Terry Tran-Nguyen} 
\author[2,3]{Adam {Bailey}}
\author[2,3]{Barbara {Braden}}
\author[2,3]{James {East}}
\author[1]{Jens {Rittscher}}
\affil[1]{Institute of Biomedical Engineering and Big Data Institute, Old Road Campus, University of Oxford, Oxford, UK}
\affil[2]{Translational Gastroenterology Unit, Experimental Medicine Div., John Radcliffe Hospital, University of Oxford, Oxford, UK}
\affil[3]{Oxford NIHR Biomedical Research Centre, Oxford, UK}
\affil[4]{Universit{\'e} de Versailles St-Quentin en Yvelines, H{\^o}pital Ambroise Par{\'e}, France}
\affil[5]{Instituto Onclologico Veneto, IOV-IRCCS, Padova, Italy}
\affil[6]{CRO Centro Riferimento Oncologico IRCCS, Aviano, Italy}
\affil[7]{Computer Engineering Department, Arab Academy for Science and Technology, Alexandria, Egypt}
\affil[8]{Wellcome/EPSRC Centre for Interventional and Surgical Sciences(WEISS) and Department of Computer Science, University College London, London, UK}
\affil[9]{CRAN UMR 7039, University of Lorraine, CNRS, Nancy, France}
\affil[10]{Graduate School of Informatics, Middle East Technical University, Ankara, Turkey}
\affil[11]{Department of Artificial Intelligence and Knowledge Systems, University of W\"urzburg, Germany}
\affil[12]{School of Engineering, University of Central Lancashire, UK}
\affil[13]{Department of Computer Science, University of Oxford, UK}
\affil[14]{Mimyk Medical Simulations Pvt Ltd, Indian Institute of Science, Bengaluru, India}
\affil[15]{Indian Institute of Technology (ISM), Dhanbad, India}
\affil[16]{Medical Imaging Department, Vingroup Big Data Institute (VinBDI), Hanoi, Vietnam}
\affil[17]{EPITA Research and Development Laboratory (LRDE), F-94270 Le Kremlin-Bicêtre, France}
\affil[18]{School of Science and Technology, Middlesex University London, UK}
\affil[19]{Department of Computer Science, School of Informatics, Xiamen University, China}
\affil[20]{Dept. of Health Sciences \& Tech., Samsung Advanced Institute for Health Sciences \& Tech. (SAIHST), Sungkyunkwan University, Seoul, Republic of Korea}
\affil[21]{Claritrics India Pvt Ltd, Chennai, India}
\affil[22]{School of Medical Science and Technology, Indian Institute of Technology, Kharagpur, West Bengal, India}
\affil[23]{Shanghai Jiaotong University, Shanghai, China}
\affil[24]{Ibn Zohr University, Computer Science HIT, Agadir, Morocco}
\affil[*]{Corresponding author: sharib.ali@eng.ox.ac.uk, ali.sharib2002@gmail.com}
\title{{Deep learning for detection and segmentation of artefact and disease instances in gastrointestinal endoscopy}\footnote{Endoscopy Computer Vision Challenge (EndoCV2020)}}
\begin{abstract}
The Endoscopy Computer Vision Challenge (EndoCV) is a crowd-sourcing initiative to address eminent problems in developing reliable computer aided detection and diagnosis endoscopy systems and suggest a pathway for clinical translation of technologies. Whilst endoscopy is a widely used diagnostic and treatment tool for hollow-organs, there are several core challenges often faced by endoscopists, mainly: 1) presence of multi-class artefacts that hinder their visual interpretation, and 2) difficulty in identifying subtle precancerous precursors and cancer abnormalities. Artefacts often affect the robustness of deep learning methods applied to the gastrointestinal tract organs as they can be confused with tissue of interest. EndoCV2020 challenges are designed to address research questions in these remits. In this paper, we present a summary of methods developed by the top 17 teams and provide  an objective comparison of state-of-the-art methods and methods designed by the participants for two sub-challenges: i) artefact detection and segmentation (EAD2020), and ii) disease detection and segmentation (EDD2020). Multi-center, multi-organ, multi-class, and multi-modal clinical endoscopy datasets were compiled for both EAD2020 and EDD2020 sub-challenges. The out-of-sample generalization ability of detection algorithms was also evaluated. Whilst most teams focused on accuracy improvements, only a few methods hold credibility for clinical usability. The best performing teams provided solutions to tackle class imbalance, and variabilities in size, origin, modality and occurrences by exploring data augmentation, data fusion, and optimal class thresholding techniques. 
\end{abstract}
\begin{document}
\flushbottom
\maketitle
\section{Introduction}
\label{sec:intro}
\begin{figure}[t!h!]
    \centering
    \includegraphics[width=0.75\textwidth]{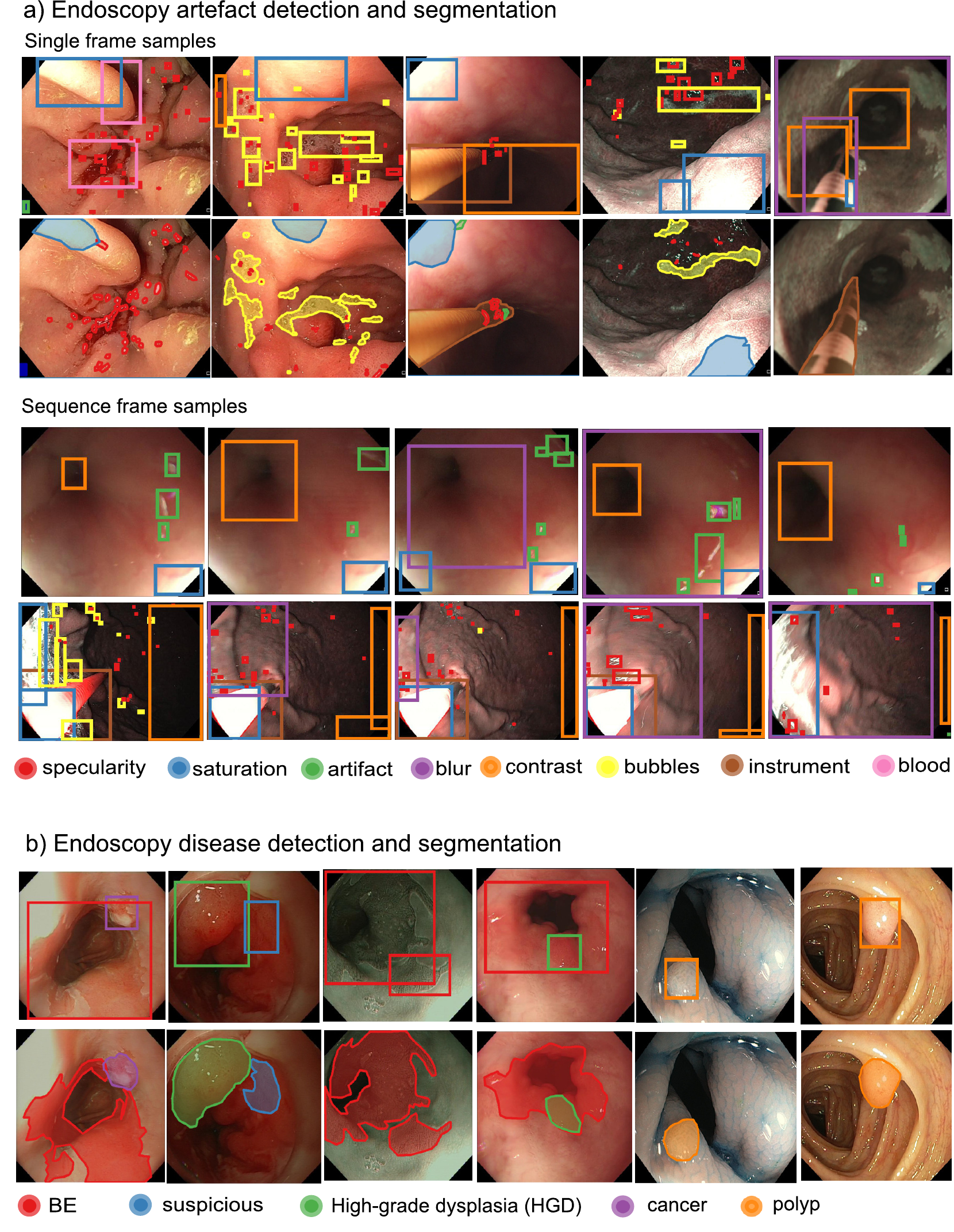}
        \caption{\textbf{EndoCV2020 train data samples.} (a) Endoscopy artefact detection and segmentation sub-challenge (EAD2020) samples. Both single frame samples (top) and sequence frames (bottom) were released. While detection annotations involve 8 classes, segmentation classes were limited to 5 distinct class instances, mostly large indefinable shapes that include specularity, saturation, imaging artefact, bubbles and instrument. It can be observed that for sequence data most artefact instances follow upto few sequential frames so it is desirable to achieve such training datasets. { 4th sample in the single frame data for segmentation shows that even though bounding boxes for detection are provided for all specular regions, some segmentation labels were missing. This shows the presence of annotator variability in the data.} (b) Endoscopy disease detection and segmentation training samples for sub-challenge EDD2020. First four samples belong to esophageal endoscopy while the last two frames were acquired during colonoscopy. It can be observed that disease classes in esophagus confuse often, mostly the patient choice here is Barrett's where clearly suspected and high-grade dysplasia appear jointly. Similarly, for colonoscopy data {protruded} polyps can easily be confused with the surrounding ridge-like openings and specular areas.~\label{fig:supplementaryaboutDatafigure2}}
 \end{figure}
Endoscopy is a widely used imaging technique for both diagnosis and treatment of patients with complications in hollow organs such as esophagus, stomach, colon, bladder, kidney and nasopharynx. During the endoscopic procedure, an endoscope, a long thin tube with a light source and a camera at its tip, is inserted into the organ cavity. The imaging procedure is usually displayed on a monitor on-the-fly and is often recorded for post analysis. Each organ imposes very specific constraints to the use of endoscopes, but the most common obstructions in all endoscopic surveillance consists of artefacts caused by motion, specularities, low contrast, bubbles, debris, bodily fluid and blood. These artefacts hinder the visual interpretation of clinical endoscopists \cite{ali2020objective}. Missed detection rates of precancerous and cancerous lesions are another limitation. Gastrointestinal (GI) cancer (especially colorectal cancer) has high mortality rates and 5-year relative survival rates for stage IIB is around 65\%~\cite{Rawla2019}. In general, the missed detection rates in endoscopic surveillance is considerably high, at over 15\%~\cite{Lee2017}. Therefore, the requirement for technology that can be effectively used in clinical settings during endoscopy imaging is necessary. 

While a dedicated endoscopic procedure is followed for each specific organ, often these procedures are very similar, in particular for the GI tract organs like the esophagus, stomach, small intestine, colon and rectum. Notably, some precancerous abnormalities such as inflammation or dysplasia and even cancer lesions in these GI organs naturally look very similar. Often automated methods are only trained for a specific abnormality, organ and imaging modality~\cite{Zhang2019PolypCNN}, whereas multiple different types of abnormalities can be present in different organs and several imaging protocols are used during endoscopy. Also, methods that are built for colonoscopy cannot be used during a gastroscopy (in the esophagus, stomach and small intestine), despite the nature and occurrence of many abnormalities being similar in these organs. Artefacts are prevalent in all endoscopy surveillance and are usually confused with lesions, which can lead to unreliable outcomes.

A pathway to develop and reliably deploy methods in clinical settings is by benchmarking methods on a curated multi-center, multi-modal, multi-organ and multi-disease dataset and through a thorough evaluation of built methods using standard imaging metrics and metrics that can test their clinical applicability, for example ranking based on accuracy, robustness and computational efficiency~\cite{ali2020objective}. Most publicly available datasets are specific to a particular organ, modality or a single abnormality class, e.g., polyp detection and segmentation challenges~\cite{bernal2017comparative,GIANA}. While dedicated organ specific challenges help to identify one particular disease type, they do not resemble the clinical workflow where the endoscopists are interested in biopsy and treatment of such abnormalities when of potential threat. For polyp class, it is required to identify different stages of polyp such as benign, dysplastic or cancer. Recently, it was shown that polyps and artefacts can be confused mostly due to specularity~\cite{Learning4mForget2020}. Artefacts are the fundamental and inevitable issue in endoscopy that often add confusion in detecting tissue abnormalities in these organs. It is therefore vital to accelerate research in identifying these classes and restore frames where possible~\cite{ALI2021101900} or reduce the false detections by adding uncertainties for such confusions~\cite{Learning4mForget2020}. {Other ways to address artefact problems in the endoscopy data is by using synthetically generated frames~\cite{FM2018:TMI,Formosa2020,incetan2020vrcaps}.~\cite{FM2018:TMI} used self-regularized transformer network that allowed to transform the real images into synthetic-like images with preserved clinically-relevant features. This allowed the authors to estimate depth in colonoscopy data robustly without being affected by adverse artefact problems. \cite{incetan2020vrcaps} demonstrated the use of a virtual active capsule environment that can simulate wide range of normal and abnormal tissue conditions such as inflated, dry and wet;  organ types and endoscopy camera designs in capsule endoscopy. This allowed to optimize the analysis software for varied real conditions.}

% TODO: connecting sentences
The Endoscopy Computer Vision Challenge (EndoCV2020)\footnote{https://endocv.grand-challenge.org} is another crowd-sourcing initiative to address fundamental problems in clinical endoscopy and consists of: 1) Endoscopy artefact detection and segmentation (EAD2020), and 2) Endoscopy disease detection and segmentation (EDD2020). EndoCV2020 releases diverse datasets that include multi-center, multi-modal, multi-organ, multi-disease/abnormality, and multi-class artefacts. Among the two sub-challenges, EAD2020 is an extended sub-challenge of EAD2019~\cite{ali2019endoscopy}, however, {unlike EAD2019 it includes both frame and sequence data with an addition of nearly 500 frames and a total of 41,832 annotations for detection task and 10,739 for segmentation task}. 

In this paper, we summarise and analyse the results of the top 17 (out of 43) teams participating in the EndoCV2020 challenge. Additionally, we benchmark these methods with the current state-of-the-art detection and segmentation methods. Each method is also evaluated for its efficacy to detect and segment multi-class instances. In addition to the standard computer vision metrics used to evaluate methods during the challenge, we perform a holistic analysis of individual methods to measure their clinical applicability. 
%Code of the participants methods are available in their github.
%%%%%
%
\section{Related work}
With the advancements in deep learning for computer vision, object detection and segmentation algorithms have shown rapid development in recent years. This is due to the hidden feature representations provided by Convolutional Neutral Networks (CNNs) that show significant improvement over hand-crafted features. CNN-based methods quickly gained the attention of the Medical Imaging community and are now widely used for automating the diagnosis and treatment for a range of imaging modalities, e.g. radiographs, CT, MRI, and endoscopy imaging. 
Below we present an overview of the recent deep learning-based object detection and segmentation techniques and discuss the related work in the context to medical image analysis with a particular focus on endoscopy imaging applications.

\subsection{Detection and localization}
Object detection and localization refers to determining the instances of an object (from a list of predefined object categories) that exist in an image. Object detection approaches can be broadly divided into three categories: single-stage, multi-stage and anchor-free detectors. A brief survey of these is presented below.
\paragraph{Single-stage detectors}
% TODO => Gorkem
% first medical imaging
Single-stage networks perform a single pass on the data and incorporate anchor boxes to tackle multiple object detection on the same image grid such as in YOLO-v2~\cite{redmon2016you}. Similarly, Liu et al. \cite{liu2016ssd} proposed the Single Shot MultiBox Detector (SSD) with additional layers to allow detection of multiple scales and aspect ratios. RetinaNet was introduced by Lin et al. \cite{lin2017focal} where the authors introduced focal loss that puts the focus on the sparse hard examples enabling a boost in performance and speed. 

%Cao et al. \cite{cao2017breast} used several object detectors to detect tumors in breast ultrasound images and they reported that SSD outperformed other methods in terms of both precision and recall. Li et al. \cite{li2017detection} improved SSD by stacking multi-level and multi-resolution features to detect pulmonary nodules from lung CT images. Lemay et al. \cite{lemay2019kidney} used YOLOv3 and SSD to localize kidney on 2D and 3D CT. 
The domain of Gastroenterology has started to benefit from the success of single-stage object detectors. Wang et al. \cite{wang2018development} proposed a model that is based on SegNet \cite{badrinarayanan2017segnet} architecture to detect polyps during colonoscopy. Urban et al. \cite{urban2018deep} used YOLO to detect polyps from colonoscopy images in real-time. Horie et al. \cite{horie2019diagnostic} used SSD to detect superficial and advanced esophagal cancer. RetinaNet was the most popular detector in the first EAD challenge held in 2019. RetinaNet detector with focal loss was used by some top performing teams~\cite{Maxime2019,Oksuz2019}
\paragraph{Multi-stage detectors}
% first medical imaging
Multi-stage detectors use a region proposal network to find regions of interest for objects and then a classifier to refine the search to get the final predictions. A two-stage architecture R-CNN using the classical region proposal method was proposed by Girshick et al.~\cite{girshick2014rich} whose speed was improved later by integrating an end-to-end trainable region proposal network (RPN), widely known as Faster R-CNN~\cite{ren2015faster}. Due to the high precision of the  Faster R-CNN, its architecture has become the base for many successful models in the object detection and segmentation domains, such as  Cascade R-CNN~\cite{cai2018cascade} and Mask R-CNN \cite{he2017mask}. Although these two-stage networks have shown successful results on public datasets such as Pascal VOC~\cite{pascal-voc-2012} and COCO~\cite{lin2014microsoft}, they are slow compared to the single-stage object detectors due to their region proposal mechanism.

%
%Ribli et al. \cite{ribli2018detecting} proposed a CAD system based on Faster R-CNN framework to detect and classify lesions in mammograms. Zhu et al. \cite{3DLung2018} proposed a 3D Faster R-CNN framework with an internal U-Net like architecture to detect lung nodules from CT Scans. 
In the field of Gastroenterology, Yamada et al. \cite{yamada2019development} used Faster R-CNN with VGG16 as the backbone to detect challenging lesions which are generally missed by colonoscopy procedures. Their reported prediction speed was not suitable for real-time examination.  Shin et.al. \cite{younghak2017} detected Polyps using the Fast R-CNN architecture with a region proposal network and an inception ResNet backbone. The two-stage detectors tend to yield better results than their single-stage contemporaries and have performed better at medical image analysis challenges. In the EAD2019 challenge, the top performing team~\cite{Suhui2019} used a Cascade R-CNN with a {feature pyramid network (FPN)} module and a ResNet backbone. Similarly, Pengyi Zhang et.al. \cite{Pengyi2019} who used Mask aided R-CNN with an ensemble of different ResNet backbones finished second. 
%Other top-performing models were based on Faster R-CNN and modifications to the Cascade R-CNN.  

\paragraph{Anchor-free detectors}
% TODO ==> Adrian Krenzer
% GIve an intro to Anchor-Free detectors
A newly emerging detector type are the anchor-free detectors. Single and multi-stage detectors rely on the presence of anchors. Anchor free architectures claim to detect objects while skipping the process of anchor definition. They rely on different geometrical characteristics like the center or corner points of objects \cite{law2018cornernet, duan2019centernet}. Duan et al. \cite{duan2019centernet} utilized the upper left and lower right corner to mark an object. The authors used classical backbones to generate a heatmap from the  feature map showing potential spots of the object corners. A corner pooling technique was then used to create the classic bounding box of object detection. Zhou et al. \cite{zhou2019bottom} used a similar approach but instead they used a single point as the center of the bounding box. 

Because of real-time dependencies in medical applications like the detection of polyps which have to be removed directly \cite{wang2019afp}, anchor-free detectors are receiving more attention. Wang et al. \cite{wang2019afp} designed an anchor-free automatic polyp detector which achieved the state-of-the-art results while maintaining real-time applicability. Liu et al. \cite{liu2020anchor} showed an anchor-free detector with state-of-the-art performance while maintaining real-time performance. 
\subsection{Semantic segmentation}
Semantic segmentation involves pixel-level partitioning of an image into multiple segments where each segment represents a pre-defined object or scene category.  Based on the success of deep learning approaches on medical imaging data for segmentation, we can divide these approaches broadly into the following groups:
% 
%TODO Check for gastroenterology based literature
\paragraph{Models based on fully convolutional networks} 
Fully Convolutional Network (FCN) architectures include only convolutional layers that enable them to take any arbitrary size input image to output a segmentation mask of the same size. These models are mostly based on the architecture developed by Long et al.~\cite{long2015fully} for semantic image segmentation. 

Sun et al. \cite{sun2017automatic} proposed a multi-channel FCN (MC-FCN) to segment liver tumors from multi-phase contrast-enhanced CT images. Kaul et al. \cite{kaul2019focusnet} proposed FocusNet for skin cancer and lung lesion segmentation. A benchmark study for polyp segmentation using FCNs was conducted by \cite{Junfeng2017}. Similarly, Patrick et al.~\cite{PatrickSPIE2017} used FCN architecture with VGG backbone for a polyp segmentation task. The same group explored integration of depth information to improve segmentation accuracy in their FCN-based model~\cite{Patrick2018FCND}.  

%The authors used an attention mechanism with convolutional neural networks and the feature maps generated by a different convolutional autoencoder. 
%Milletari et al. \cite{milletari2016v} developed a model called V-net for 3D image segmentation of  MRI volumes of prostrate.
%
%TODO add Soumya's work -- UNet, Instrument segmentation of Debesh, 
\paragraph{Models based on encoder-decoder architecture} 
U-Net~\cite{ronneberger2015u}, an encoder-decoder architecture, has become widely popular in medical image analysis community. U-Net based models have shown tremendous success, from cell segmentation~\cite{Falk2019} to liver tumor segmentation~\cite{chlebus2017neural} and beyond~\cite{sevastopolsky2017optic,norman2018use}.

%Chlebus et al.~\cite{chlebus2017neural} designed a 28-layer U-Net architecture which came in second place in the ISBI 2017 challenge for liver tumor segmentation. Sevastopolsky et al.~\cite{sevastopolsky2017optic} applied U-Net to segment the optic disc and optic cup in retinal fundus images for glaucoma diagnosis. Norman et al.~\cite{norman2018use} used U-Net to segment cartilage and meniscus from knee MRI data. Chen et al. \cite{chen2016dcan} developed a model called DCAN (Deep Contour Aware Network) based on U-Net architecture, which won the MICCAI Gland Segmentation Challenge. 

In endoscopy imaging, U-Net-based models were used for instrument segmentation on GI endoscopy data~\cite{jhaInstrument2020}. Khan and Choo \cite{khan2019multi} developed a model based on U-Net architecture for endoscopy artefact segmentation. Bano et al.~\cite{bano2020deep} directly used U-Net architecture for segmenting placental vessels from Fetoscopy imaging. Motion induced segmentation exploiting U-Net in the framework was used to segment kidney stones in the Uteroscopy data \cite{SG2020}. 
\paragraph{Models based on pyramid-based architecture}
In both detection and segmentation tasks, a crucial part is being able to identify objects and features of varying scales and sizes. One approach to this problem is to incorporate convolutional feature maps of varying resolutions during classification, which yields information about different scales of the image, making it easier to detect both small and big objects. Such architectures are referred to as \textit{pyramid networks}. PSPNet~\cite{zhao2017pyramid} is one of such design that incorporates global context information for the task of scene parsing using a pyramid pooling module. A similar pyramid-based approach can be found in the task of object detection with Feature Pyramid Network (FPN) \cite{lin2017feature}. FPN extracts feature maps on a per-resolution-basis from the two bottom-up and top-down pathways of a pretrained architecture. The output maps can then be upsampled and concatenated to output a segmentation map \cite{selim2018FPN}.

%Men et al. \cite{Men2018} adopted a spatial pyramid pooling module in a cascade CNN for the segmentation of rectal tumors in CT images. 

Guo et al. \cite{Guo2019} used PSPNet as part of an ensemble model including a U-Net and SegNet architecture for the task of automated polyp segmentation in colonoscopy images. 
%Similarly, FPN has also been successfully applied on medical data. Solovyev et al. \cite{Solovyev2020} used a modified Bayesian FPN for multi-label segmentation of chest X-ray. Several applications of FPN can be found in literature for lesion detection, for e.g.,  modified FPN to focus on the detection of retinopathy in fundus images \cite{Chen2019minilesions}, a multi-view FPN for universal lesion detection in CT images \cite{Li2019MVPNetMF} and an attention-based FPN with a novel Multi-Scale Booster for lesion detection in CT images \cite{Shao2019AttentiveCL}. 
Jia et al. \cite{Jia2020} trained a two-stage polyp detector named PLPNet which utilizes FPN for multiscale feature representation using both CVC-ColonDB~\cite{bernal2012towards} and CVC-ClinicDB~\cite{bernal2015wm}. Their experimental results show that PLPNet outperforms other architectures in most regions on CVC-612 dataset~\cite{bernal2015wm} and performs similarly on the ETIS dataset~\cite{silva2014toward}. Zhang and Xie \cite{zhang2019detection} utilized an FPN combined with a Cascade R-CNN for artefact detection in endoscopic images.
\paragraph{Models based on dilated convolution architecture}
% %
One of the challenges in the construction of semantic segmentation networks is to effectively control the size of the receptive field, providing adequate contextual information for pixel-level decisions while, at the same time, maintaining high spatial resolution and computational efficiency.~The~{\textit{dilated} or \textit{atrous}} convolution was proposed to address these challenges~\cite{yu2015multi}.~Chen at al. \cite{chen2018encoder} proposed a family of very effective semantic segmentation architectures, collectively named DeepLab (also an \textit{encoder-decoder} network), all using the dilated convolution. DeepLabv3+ uses atrous kernels within the spatial pyramid pooling (ASPP) module and depth-wise separable convolution to improve the computational efficiency.
%and adopts the encoder-decoder architecture to better utilize multi-scale contextual information. 

%DeepLabv3+ was reported \cite{chen2014semantic,chen2017deeplab,chen2017rethinking,chen2018encoder} to outperform the other state of the art techniques, including PSPNet, on the PASCAL VOC 2012 \cite{pascal-voc-2012} and Cityscapes datasets\cite{cordts2016cityscapes}. 

%Perone et al. \cite{perone2018spinal} employed DeepLabv3 network to segment human spinal cord gray matter images and achieved better performance than existing methods with fewer model parameters. Wen et al. \cite{wen2019segmentation} developed an attention model based on DeepLabv3+ architecture for segmentation of kidney lesions. 

%Men et al. \cite{Men2018} used ASPP to learn multi-scale information about rectal tumors in MRI and CT images, with the dilation rate in ASPP reduced to learn more features in the most valid regions.  For cell counting and localization, Rad et al. \cite{rad2019cell} added a feature extractor constructed by cascaded dilated convolution after the backbone and obtained denser features by fusing their respective outputs. A similar method has been proposed by Anthimopoulos et al.~\cite{anthimopoulos2018semantic}. In their lung tissue segmentation task, the entire encoder is constructed using cascaded dilated kernels. 

Guo et al.~\cite{guo2020polyp} proposed a fully convolutional network based on atrous kernels to segment polyps in endoscopy images, with their network winning the GIANA 2017 challenge \cite{GIANA}. Nguyen et al. \cite{nguyen2020contour} augmented DeepLabv3+ architecture, showing its favourable performance when compared with other state-of-the-art methods on the CVC-ClinicDB~\cite{bernal2015wm} and ETIS-Larib~\cite{silva2014toward} datasets. Ali et at.~\cite{Ali20203D} used DeepLabv3+ with ResNet50 backbone to segment Barrett's area from esophageal endoscopy data. Yang and Cheng~\cite{yangendoscopic} developed a model based on DeepLabv3+ for multi-class artefact segmentation used with different backbone architectures. 
\subsection{Endoscopy computer vision challenges}
{Biomedical challenges allow to  set-up  a  benchmark  for  different computer  vision  methods. Several sub-challenge categories for the development of automated methods for wide-range of problems in endoscopy including surgical instrument segmentation~\cite{RO2020101920}, robotic scene segmentation~\cite{allan20202018}, and computer aided detection and segmentation for polyps~\cite{bernal2017comparative,bernal2018polyp} and Barrett's cancer detection\footnote{https://endovissub-barrett.grand-challenge.org} have been initiated under MICCAI EndoVis challenge\footnote{https://endovis.grand-challenge.org}. Endoscopy artefact detection (EAD2019) is another challenge which was first initiated in 2019 and  launched in conjunction with IEEE  International Symposium on Biomedical Imaging (ISBI) 2019~\cite{ali2020objective}.}
\section{The EndoCV challenge: Dataset, evaluation and submission}
In this section, we present the dataset compiled for the EndoCV2020 challenge, the protocol used to obtain the ground truth for this data, evaluation metrics that were defined to assess participants methods and a brief summary on the challenge setup and ranking procedure.
\subsection{Dataset and challenge tasks}{\label{Sec:datasets}}
\begin{figure}[t!h!]
    \centering
    \includegraphics[width= 0.8\textwidth]{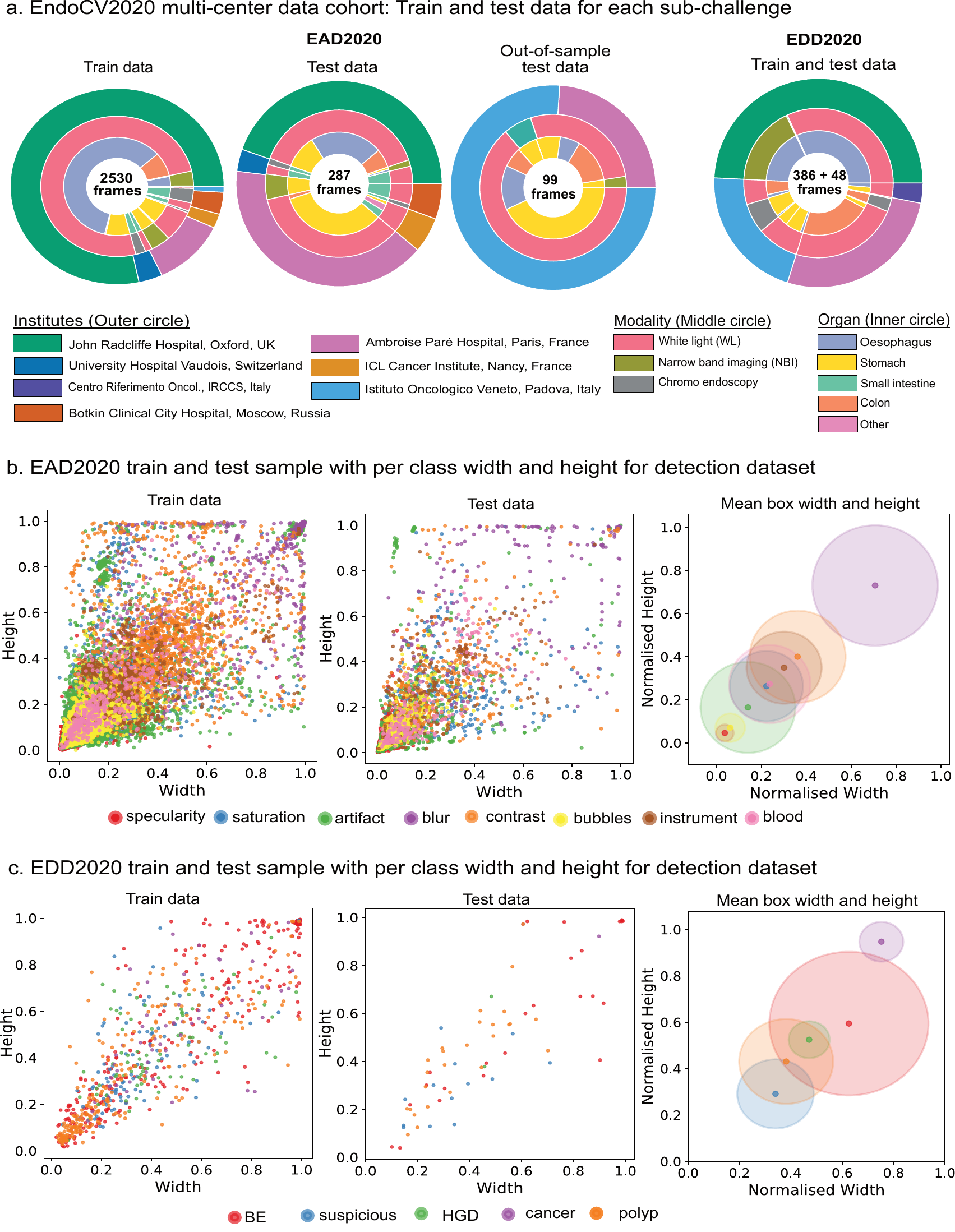}
    \caption{\textbf{Endoscopy computer vision EndoCV2020 challenge dataset details.} (a) Multi-center, multi-modality and multi-organ dataset for EAD and EDD sub-challenges. For EAD2020, 2532 frames with 8 class bounding boxes for the detection task out-of which 573 included ground truth masks for segmentation task were provided. Participants were assessed on 317 frames for detection and 162 frames for segmentation tasks. An additional 99 frames were used to test out-of-sample generalization task for EAD sub-challenge. While EDD2020 consisted of 384 train samples and 43 test samples for 5 disease classes. (b-c) The distribution of 8 artefact classes of EAD and 5 disease classes of EDD  w.r.t. their size compared to their height and width of image is provided. Each class size variability is also shown on right as blobs with mean at center and radius as standard deviation.~\label{fig:figure2}}
\end{figure}
%\subsection*{Overview of the EndoCV Challenge}
The EndoCV2020 challenge consists of two sub-challenges critical in clinical endoscopy. 
The EAD2020\footnote{https://ead2020.grand-challenge.org} sub-challenge comprises of diverse endoscopy video frames collected from seven institutions worldwide, including three different modalities and five different human organs (see Figure~\ref{fig:figure2}).  Endoscopy video frames were annotated for detection and localization of eight different artefact class occurrences identified by clinical experts in the challenge team. These include specularity, saturation, misc. artefacts,  blur, contrast, bubbles, instrument and blood. {A total of 280 patient videos from multiple organs and institutions have been used for curating this dataset.} Over 45,478 annotations were performed for this challenge on both single frame and sequence video data. Example annotations are shown in Figure~\ref{fig:supplementaryaboutDatafigure2}. Training data for the detection task consisted of total 2,531 frames with 31,069 bounding boxes while {643 frames with 7,511  binary masks} were released for the segmentation task (except for blur, blood and contrast). { The sequence data were sampled by manually observing the amount of changes in artefact categories in the selected sequence. Sequences were required to change from large areas of artefacts to small or no artefact frames and vice versa mimicking natural occurrence in endoscopic procedures. Sequence data for training included 5 sequences {(232 frames)} for detection and 2 sequences (70 frames) for semantic segmentation tasks sampled from 3 videos of 3 different patients.}  {For the test set, two sequence (80 frames) for detection task were used from 2 independent patient videos.} As observed in Figure~\ref{fig:figure2}, due to the nature of occurrence of various artefact classes, the proportion of annotations for each class is different (Figure~\ref{fig:dataset_samples}). However, the proportion of training and test samples per-class were matched in the test data (also see Table~\ref{tab:dataBreakdown}). 
%%%%%%%%%%%%%%%
\begin{table*}
\centering
\caption{\textbf{Breakdown of  data:} Number of samples and annotations released for EndoCV2020 challenge. \label{tab:dataBreakdown}}
\begin{tabular}{l|l|c|c|c|c|c}
\hline
\multicolumn{1}{c|}{\multirow{2}{*}{\textbf{EndoCV}}} & \multicolumn{1}{c|}{\multirow{2}{*}{\textbf{Tasks}}}& \multicolumn{1}{|c|}{\multirow{2}{*}{\textbf{\# of classes}}} & \multicolumn{2}{c}{\textbf{\# of frames}} & \multicolumn{2}{|c}{\textbf{\# of annotations}} \\ \cline{4-7} 
\multicolumn{1}{c|}{} & \multicolumn{1}{|c}{} &\multicolumn{1}{|c}{} &\multicolumn{1}{|c}{\textbf{Train}}  & \textbf{Test}& \textbf{Train}          & \textbf{Test}         \\ \hline
\multirow{3}{*}{EAD2020}& Detection task & 8 & \multicolumn{1}{l|}{\begin{tabular}[c]{@{}l@{}}single: 2299\\ seq.: 232\end{tabular}} & \multicolumn{1}{l|}{\begin{tabular}[c]{@{}l@{}}single: 237\\ seq.: 80\end{tabular}} & 31069                   & 7750                  \\ \cline{2-7} 
& Segmentation task & 5  & {643}  & 162  & {7511} & {3228}\\ \cline{2-7} 
& Generalization task & 8  & na & 99 & na  & 3013 \\ \hline
\multirow{2}{*}{EDD2020} & Detection task & 5 & 386 & 43 & 749 & 68 \\ \cline{2-7} 
& Segmentation task & 5 & 386 & 43 & 749 & 68 \\ \hline
\end{tabular}
\end{table*}
\begin{figure}[t!]
    \centering
    \includegraphics[width= 1.0\textwidth]{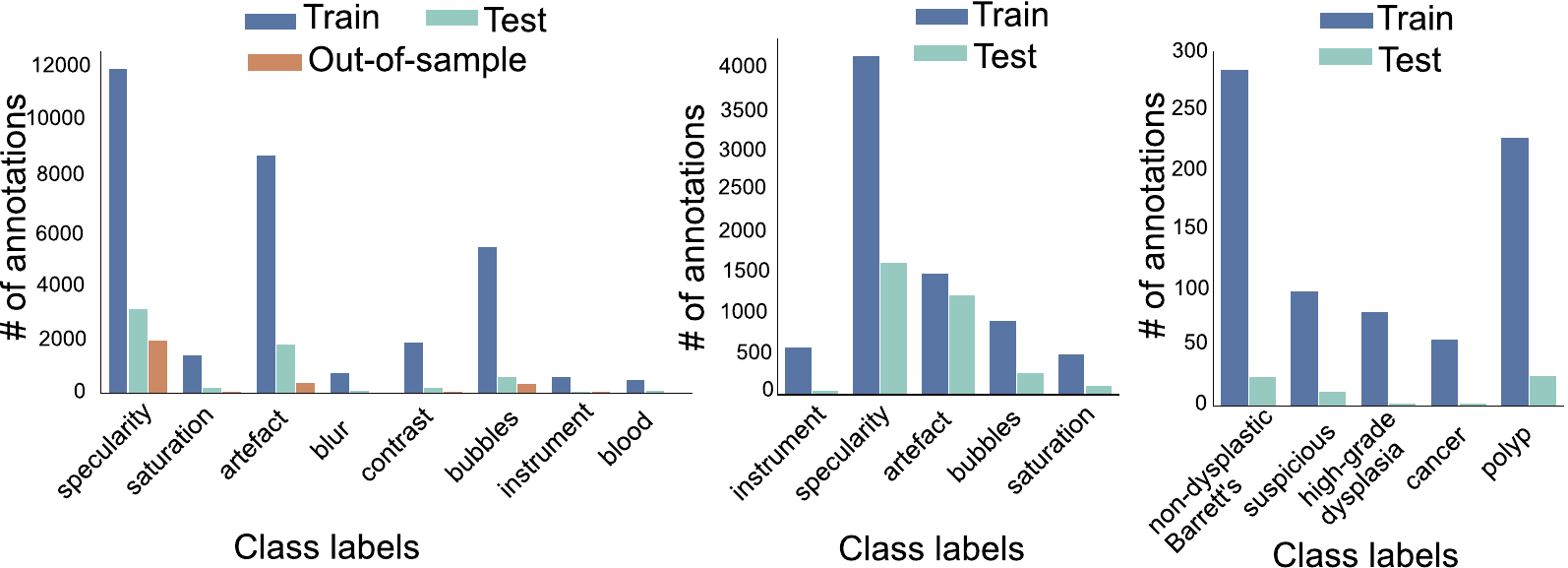}
    \caption{\textbf{EndoCV2020 train and test per-class sample proportion}: Train and test annotations for sub-challenge on artefact (A,B) and disease (C) detection and segmentation for each class label.}
    \label{fig:dataset_samples}
\end{figure}
%%%%s
%

Separately, EDD2020\footnote{https://edd2020.grand-challenge.org} is a new disease detection and segmentation sub-challenge that consists of five disease categories~\cite{EDD2020}. The provided training set consisted of total 385 video frames {comprising of 137 different patients used in this study} with a total of 817 individual annotations. The annotations included non-dysplastic Barrett's esophagus (NDBE), suspicious, high-grade dysplasia (HGD), cancer, and polyp categories (also see Figure~\ref{fig:supplementaryaboutDatafigure2}). These disease classes were from three different endoscopic modalities (white light, narrow-band imaging, and chromoendoscopy) acquired from four different clinical centers, investigating four different GI organs. %Are the methods developed in this work equally applicable to colonoscopy or are they limited to upper GI?
{By including varied range of endoscopy data acquired from multiple organs like GI tract and liver in EAD sub-challenge and both upper and lower GI tract data for EDD sub-challenge, EndoCV2020 challenge aimed at developing more general methods that can potentially be applied in different endoscopy routine procedures independent to organ type.} To our knowledge, this is the first comprehensive dataset for the multi-class detection and segmentation tasks. More details on the dataset are provided in Figure~\ref{fig:figure2}. The detailed breakdown of training set and test set for each specific task is provided in Table~\ref{tab:dataBreakdown}.

\begin{figure}[t!]
    \centering
    \includegraphics[width= 0.9\textwidth]{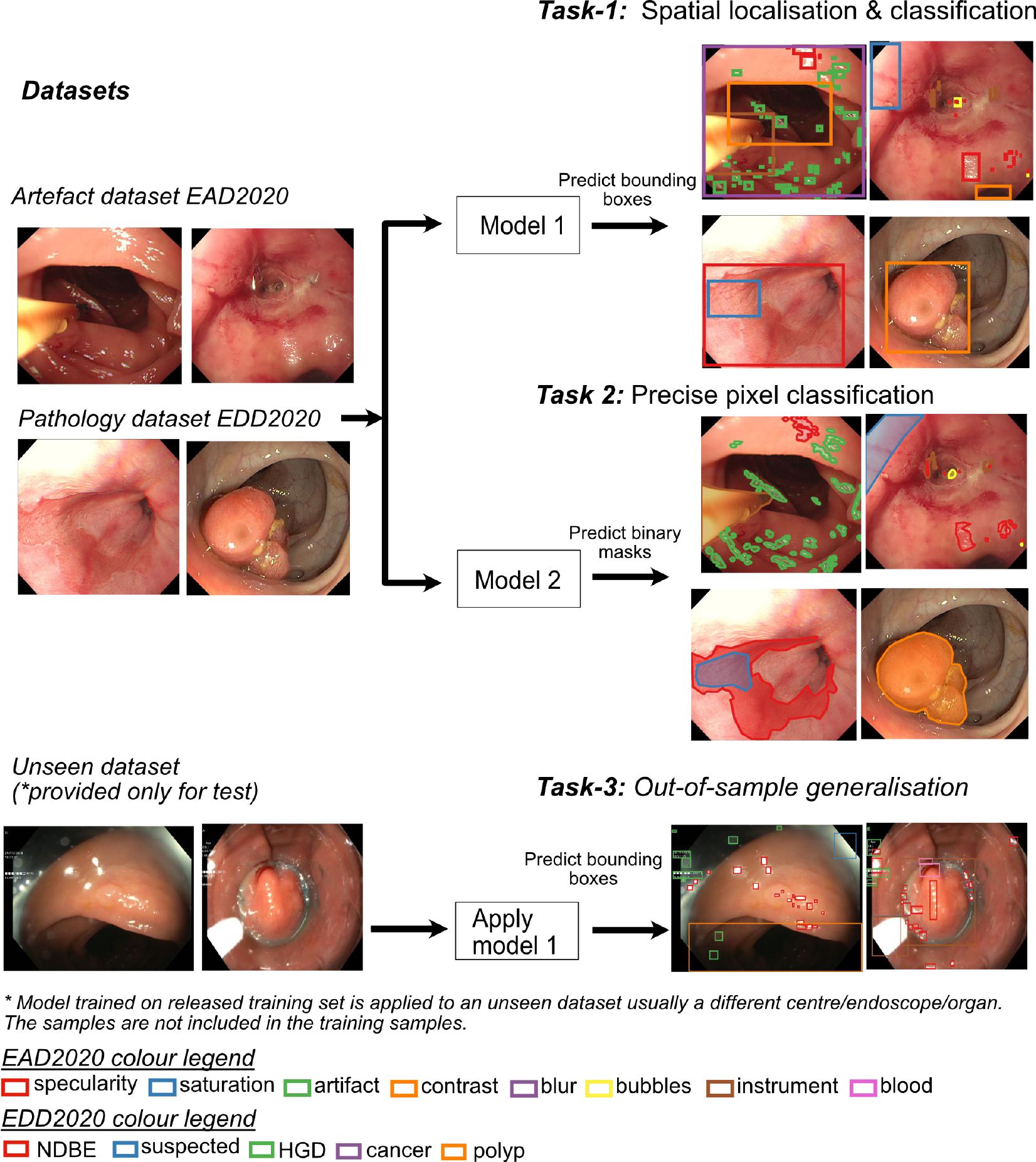}
    \caption{\textbf{EndoCV2020 challenge task descriptions for each sub-challenge.} The three tasks of the EndoCV2020 challenge includes: (a) The ``detection'' task aimed at the coarse localization and classification. Given an input image (left) a detection model (middle) outputs the artefact/disease class and coordinates of the containing bounding box. (b) The ``segmentation'' task is aimed at precise delineation of artefact/disease object boundaries. The model predicts binary output images denoting the presence (‘1’) or absence (‘0’) of each class. (c) The ``out-of-sample generalization'' task is aimed at assessing the ability of a model trained on different dataset to generalize on an unseen dataset usually coming from a different center.}
    \label{fig:endoCV_scopes}
\end{figure}

% =========
EndoCV2020 posed three specific challenge tasks (see Figure~\ref{fig:endoCV_scopes}) that included: 1) detection and localization task, 2) semantic segmentation task and 3) out-of-sample generalization task. For detection and generalization tasks, participants were provided with both frame label annotations for single and sequence images for the EAD2020 challenge while only single frames were released for EDD2020. The generalization task was only evaluated for the EAD2020 and only consisted of test data from an unseen institution that was not present in any training set. It is to be noted that test samples for all other tasks were taken from different patients as well even though they were collected from the same centers as that in the training set.
%
% About Challenge participations
EAD2020 attracted nearly 700 participants with 29 teams on the leaderboard and EDD2020 recorded nearly 550 participants with 14 teams on the leaderboard. Participation was permitted in either one or both sub-challenges. 
% Dataset availability
{Both challenge datasets are publicly available for research and education. EAD2020 challenge data is available at Mendeley Data ({\url{10.17632/c7fjbxcgj9.3}}) and EDD2020 dataset is available at IEEE dataPort ({\url{http://dx.doi.org/10.21227/f8xg-wb80}})}.

% Sample images for artefact detection
\begin{table*}[t!]
{
\small{
\begin{tabular}{l|l|l|l}
\hline
\textbf{Centers}                              & \textbf{System info.}                & \textbf{Ethical approval} & \textbf{Patient consenting type} \\ \hline
John Radcliffe Hospital, Oxford, UK           & Olympus GIF-H260Z, & REC Ref: 16/YH/0247       & Universal consent                \\ 
         & EVIS Lucera CV260 &    & \\ \hline
Ambroise Par\'{e} Hospital, Paris, France         & Olympus Exera 195                    & No. IDRCB: 2019-A01602-55  & Endospectral study   \\ \hline
Istituto Oncologico Veneto, Padova, Italy     & Olympus endoscope H190               & NA                        & Generic patients consent         \\ \hline
Centro Riferimento Oncologico, IRCCS, Italy   & Olympus VG-165, CV180, H185          & NA                        & Generic patients consent         \\ \hline
ICL, Cancer Institute, Nancy, France          & Karl Storz 27005BA                   & NA                        & Generic patients consent         \\ \hline
University Hospital Vaudois, Switzerland      & NA (flexible cystoscopy)             & NA                        & Generic patients consent         \\ \hline
Botkin Clinical City Hospital, Moscow, Russia & BioSpec                              & NA                        & Generic patients consent         \\ \hline
\end{tabular}
}
}
\caption{\textbf{Data collection information for each center:} Data acquisition system and patient consenting information. \label{tab:patientConsentingInfo}}
\end{table*}
{
\subsubsection{Ethical and privacy aspects of the data}
Data for EAD2020 were collected from 7 different centers while for EDD2020 were from 4 centers. Each center was responsible for handling the ethical, legal and privacy of the relevant data sent to the challenge organizers. The data collection from each center included either two or all essential steps described below:
\begin{itemize}
\item[1.] Patient consenting procedure at the home institution (required)
\item[2.] Review of the data collection plan by a local medical ethics committee or an institutional review board
\item[3.] Anonymization of the video or image frames {(including demographic information)} prior to sending to the organizers (required)
\end{itemize}

\noindent Table~\ref{tab:patientConsentingInfo} illustrates the ethical and legal processes fulfilled by each center along with the endoscopy equipment and recorders used for the data collected for this challenge.}
\subsubsection{Annotation protocol}
A team of two clinical experts and {one post-doctoral researcher} determined the class labels for the artefact detection challenge while for the disease detection challenge we consulted with {four senior Gastroenterologists (over 20 years experience)} regarding the class labels in the GI tract endoscopy. For each sub-challenge senior Gastroenterologists sampled the video frames from a small sub-set of video data collected from various institutions and multi-patient data cohort (see Figure~\ref{fig:figure2}). These frames were then taken as reference to produce bounding box annotations for the remaining train-test dataset by four experienced postdoctoral fellows. Finally, further validation by three clinical endoscopists independently was carried out to assure the reference standard. The ground-truth labels were randomly sampled (1 per 20 frames) during this process. However, after the completion of this phase the entire annotation was discussed and reviewed together with the team of senior Gastroenterologists. Priority was given to indecisive frame annotations to have a collective opinion from experts.
% Annotation Protocols:
%
Following general annotation strategies were used by clinical experts and researchers:
\begin{itemize}
    \item  For the same region, multiple boxes (for detection/generalization) or pixel-wise delineation (for semantic segmentation) were performed if the region belonged to more than 1 class 
    \item The minimal box sizes were used to describe the class region and similarly possible small annotation areas for semantic segmentation were merged instead of having multiple small boxes/regions
    \item Each class type was determined to be distinctive and general across all datasets
\end{itemize}

{
For EAD dataset, defined class categories used included below descriptions~\cite{ALI2021101900}. Related samples are presented in Fig.~\ref{fig:supplementaryaboutDatafigure2} (a).
\begin{enumerate}
\item blur $\rightarrow$ fast camera motion 
\item bubbles	$\rightarrow$ a thin film of liquid with air that distorts tissue appearance 
\item specularity $\rightarrow$	mirror-like reflection 
\item saturation	$\rightarrow$ overexposed bright pixel areas
\item contrast $\rightarrow$ low contrast areas from underexposure  
\item misc. artifact $\rightarrow$ chromatic aberration, debris etc. 
\item instrument $\rightarrow$ biopsy or any other instrument
\item blood $\rightarrow$ flow of red colored liquid due to biopsy or surgery
\end{enumerate}
% cite clinical papers here
For EDD dataset, both upper-GI (gastroscopy) and lower-GI  (colonoscopy) data were used with below defined class categories (please refer to the samples in Fig.~\ref{fig:supplementaryaboutDatafigure2} (b)):
\begin{enumerate}
\item NDBE or BE $\rightarrow$ non-dysplastic Barrett's esophagus determined by a squamo-columnar junction above the gastric fold in the esophagus~\cite{Eluri:BE2017} 
\item HDG $\rightarrow$ high-grade dysplasia or early adenocarcinoma determined by irregular mucosal appearance~\cite{Wang:BE2012} 
\item suspected $\rightarrow$ aka low-grade dysplasia, an early sign of pathology~\cite{Eluri:BE2017} 
\item cancer $\rightarrow$ abnormal growth~\cite{Boland2005}
\item polyp $\rightarrow$ abnormal protrusion of the mucosa~\cite{Williams:Colorectal2013} 
%https://www.acpgbi.org.uk/content/uploads/Management-of-the-Malignant-Colorectal-Polyp.pdf
\end{enumerate}
}
{For the annotations of disease classes, pathology reports were also used to validate the class category for non-dysplastic Barrett's esophagus (BE), high-grade dysplasia (HGD), suspected (dysplasia or low-grade dysplasia), and cancer categories. That is, expert annotations (three senior gastroenterologists) were taken and supported with the pathology report for most disease categories including some indecisive cases. However, for the polyp class, both the protruded and flat polyps were marked by two experienced post-doctoral researchers and checked by a senior lower-GI specialist (no further categorization based on pathology report was done except for cancer cases).}

%As the reviewer pointed out regarding the observer variability in both pathology and endoscopy, we relied mostly on over 20 years experienced four gastroenterologists to curate this dataset. Additionally, to resolve the conflict in some frame annotations, we decided to use the pathology information logged into the system. Since, this is a retrospective study, it is difficult to rectify any difference that might have come from pathology. This is the reason why we used clinical expert understanding for most of our annotations. 

%
\subsection{Evaluation metrics}{\label{Sec:evaluation_metrics}}
The challenge problems fall into three distinct categories. For each there already exist well-defined evaluation metrics used by the wider imaging community which we use for evaluation here. Codes related to all evaluation metrics used in this challenge are also available online\footnote{https://github.com/sharibox/EndoCV2020}.
\subsubsection{Spatial localization and classification task}
Metrics used for multi-class disease detection: 
\begin{itemize}
    \item IoU - intersection over union: This metric measures the overlap between two bounding boxes $A$ and $B$, where A is segmented region and B is annotated GT. It is evaluated as the ratio between the overlapped area $A\cap B$ over the total area $A\cup B$ occupied by the two boxes:
    \begin{equation}
        \mathrm{IoU} = \frac{A\cap B}{A\cup B}
    \end{equation}
    where $\cap$, $\cup$ denote the intersection and union respectively. In terms of numbers of true positives (TP), false positives (FP) and false negatives (FN), IoU (aka {Jaccard JC}) can be defined as:
    \begin{equation}
        IoU/JC = \frac{TP}{TP+FP+FN}
    \end{equation}
    
    \item mAP - mean average precision: mAP of detected class instances is evaluated based on precision (p) defined as $p={\frac {TP}{TP+FP}}$ and recall (r) as $r={\frac {TP}{TP+FN}}$. {This} metric measures the ability of an object detector to accurately retrieve all instances of the ground truth bounding boxes. Average precision (AP) is computed as the Area Under Curve (AUC) of the precision-recall curve of detection sampled at all unique recall values $(r_1, r_2,...)$ whenever the maximum precision value drops:
    \begin{equation}
        \mathrm{AP} = \sum_n{\left\{\left(r_{n+1}-r_{n}\right)p_{\mathrm{interp}}(r_{n+1})\right\}}, 
    \end{equation}
    % \begin{equation}

    % \end{equation}
    with $p_{\mathrm{interp}}(r_{n+1}) =\underset{\tilde{r}\ge r_{n+1}}{\max}p(\tilde{r})$. Here, $p(r_n)$ denotes the precision value at a given recall value. This definition ensures monotonically decreasing precision. The mAP is the mean of AP over all $N$ classes given as
    \begin{equation}
        \mathrm{mAP} = \frac{1}{N}\sum_{i=0}^{N}{\mathrm{AP}_i}
    \end{equation}

This definition was popularised in the PASCAL VOC challenge~\cite{pascal-voc-2012}. 
    %The calculation is illustrated in Fig. \ref{fig:detection_metrics_illustration}. 
    The final mAP (mAP$_d$) was computed as an average mAPs for IoU from 0.25 to 0.75 with a step-size of 0.05 which means an average over 11 IoU levels is used for 5 categories in the competition (mAP $@[.25:.05:.75]$ ).
\end{itemize}
Participants were finally ranked on a final mean score $(\mathrm{score_d})$, a weighted score of mAP and IoU represented as:
\begin{equation}{\label{eq:score_d}}
\mathrm{score_d} = 0.6 \times \mathrm{mAP_d} + 0.4 \times \mathrm{IoU_d}
\end{equation}
Standard deviation between the computed mAPs ($\pm \sigma_{score_d}$) are taken into account when the participants have the same $\mathrm{score_d}$. Scores on both single frame data and sequence data were first separately computed and then averaged to get the final ${score_{d}}$ of the detection task.
\subsubsection{Segmentation task}
Metrics widely used for multi-class semantic segmentation of disease classes have been used for scoring semantic segmentation. The final semantic score ${score_s}$ comprises of an average score of $F_1$-score (Dice Coefficient, DSC), $F_2$-score, precision (PPV), recall (Rec) and accuracy (Acc).

\textbf{Precision, recall, F$_\beta$-scores:}
These measures evaluate the fraction of correctly predicted instances. Given a number of true instances $\#\mathrm{GT}$ (ground-truth bounding boxes or pixels in image segmentation) and number of predicted instances $\#\mathrm{Pred}$ by a method, precision is the fraction of predicted instances that were correctly found, $PPV=\frac{\#\mathrm{TP}}{\#\mathrm{Pred.}}$ where $\#\mathrm{TP}$ denotes number of true positives and recall is the fraction of ground-truth instances that were correctly predicted, $Rec=\frac{\#\mathrm{TP}}{\#\mathrm{GT}}$. Ideally, the best methods should have jointly high precision and recall. $\mathrm{F_\beta}$-scores gives a single score to capture this desirability through a weighted ($\beta$) harmonic means of precision and recall, $\mathrm{F_\beta} = (1 + \beta^2) \cdot \frac{{PPV} \cdot {Rec}}{(\beta^2 \cdot {PPV}) + {Rec}}$.
  
Participants are ranked based on the value of their semantic performance score given by:
\begin{equation}{\label{eq:score_s}}
\mathrm{score_s} = 0.25\times  (p + r + \mathrm{F_1} + \mathrm{F_2} )
\end{equation}
Standard deviation between each of the subscores are computed and averaged to obtain the final $\pm \sigma_{score_s}$ which is used during evaluation for participants with same final semantics score. {We have also used provided accuracy of each semantic method in this paper for scientific completeness. Accuracy (Acc) can be defined as $ Acc = \frac{TP+TN}{TP+TN+FP+FN}$.}
\subsubsection{Out-of-sample generalization task}
Out-of-sample generalization of disease detection is defined as the ability of an algorithm to achieve similar performance when applied to a completely different institution data. To assess this, participants were challenged to apply their trained models on video frames that were neither included in the training nor in the test data of the other tasks. Assuming that participants applied the same trained weights, the out-of-sample generalization ability was estimated as the mean deviation between the mAP score of the detection and out-of-sample generalization test datasets of each class $i$ for deviation greater than a tolerance of $\{0.1~\times$~mAP$_\text{d}^i\}$.
\begin{align}
    \mathrm{dev_g} & = \frac{1}{N} \sum_{i}{\mathrm{dev_g}^i} \\
    \mathrm{dev_g}^i & = \begin{cases}
                            0, &\text{for } |\mathrm{mAP_d}^{i} - \mathrm{mAP_g}^{i}|/\mathrm{mAP_d}^{i} \leq 0.1 \\
                            |\mathrm{mAP_d}^{i} - \mathrm{mAP_g}^{i}|, &\text{for } |\mathrm{mAP_d}^{i} - \mathrm{mAP_g}^{i}| / \mathrm{mAP_d}^{i} > 0.1 
                         \end{cases}
\end{align}
% add details of the penalisation. 
The best algorithm should have high $\mathrm{mAP_g}$ and low $\mathrm{dev_g} (\rightarrow 0$). Participants were finally ranked using a weighted ranking \textit{score for out-of-sample generalization} as $\mathrm{R_{gen}}=1/3\cdot\mathrm{Rank({dev_g})}+2/3\cdot\mathrm{Rank({mAP_g})}$ where $\mathrm{Rank({mAP_g})}$ is the rank of a participant when sorted by $\mathrm{mAP_g}$ in ascending order.
%%%%
\subsection{Challenge setup, and ranking procedure}
%%%
The challenge proposal was submitted to the IEEE ISBI challenge organisers and was peer-reviewed by two reviewers. Upon the acceptance, the challenge website\footnote{\url{https://endocv.grand-challenge.org}} was launched on 1st November 2019. Training datasets for each sub-challenge (EAD and EDD) were first provided (via AWS amazon S3 for EAD data and IEEE data portal for EDD data\footnote{\url{https://ieee-dataport.org/competitions/endoscopy-disease-detection-and-segmentation-edd2020}}). The test data was released nearly 20 days before the leaderboard closing through a docker container set-up. A docker based online leaderboard was established separately for EAD2020\footnote{\url{https://ead2020.grand-challenge.org/evaluation/leaderboard/}} and EDD2020\footnote{\url{https://edd2020.grand-challenge.org/evaluation/leaderboard/}} where each participating team was allowed to submit a maximum of 2 submissions per day on the final test data. A wiki-page\footnote{\url{https://github.com/sharibox/EndoCV2020/wiki}} was set-up for the submission guidelines and a code repository with evaluation metrics used in the challenge was also provided\footnote{\url{https://github.com/sharibox/EndoCV2020}}.  

%Ranking protocol and justification 
For the ranking of different task categories, we used the metrics described in Section~\ref{Sec:evaluation_metrics}. The participants were able to see only the final score in the leaderboard and all other sub-scores were hidden for the final test data. This was done to avoid any class specific refinement on the released test set. Notably, the detection task was bounded by two IoU thresholds (mAP $@$ IoU thresholds $[.25:.05:.75]$) and the overall IoU scores itself. For the detection task, participants were ranked on a final weighted score of mAP and IoU (see Eq.~(\ref{eq:score_d})), while for the segmentation task, participants were ranked based on a final weighted average of DSC or F1-score, F2-score, precision and recall (see Eq.~(\ref{eq:score_s})). For the generalization task, both the mAP score gap $ \mathrm{dev_g}$ and mAP on generalization data $\mathrm{mAP_g}$ were taken into account.
\section{Method summary of the participants}
%
%\begin{landscape}
\begin{table*}
\caption{Endoscopy artefact detection and segmentation (EAD2020) method summary for top 13 teams (out-of 33 valid submissions).~\label{tab:EAD2020_method_summary}}
\centering
\begin{adjustbox}{width=1\textwidth}
\small{
\begin{tabular}{l|l|l|l|l|l|l|l|l|l|l} 
\hline\hline
\multicolumn{1}{l|}{\textbf{Team EAD2020}}                      & \multicolumn{1}{l|}{\textbf{Algorithm}}                                          & \multicolumn{1}{l|}{\textbf{Preprocessing}}                          & \multicolumn{1}{l|}{\textbf{Nature}}                                & \multicolumn{1}{l|}{\textbf{Basis-of-choice}}               & \multicolumn{1}{l|}{\textbf{Backbone}}                                   & \multicolumn{1}{l|}{\textbf{Data aug.}}                                           & \multicolumn{1}{l|}{\textbf{Pretrained}}              & \multicolumn{2}{c|}{\textbf{Computation}}                                                                                    & \multicolumn{1}{c}{\textbf{code}}         \\ 
\hline\hline
\multicolumn{8}{l|}{\begin{tabular}[c]{@{}l@{}}\textbf{}\\\textbf{Detection~} \end{tabular}}                                                     & \multicolumn{1}{l|}{\textbf{GPU}}                                 & \multicolumn{1}{l|}{\textbf{Test time}}                 &                                             \\ 
\hline
%\hline{|-|-|-|-|-|-|-|-|=|=|-|}
\begin{tabular}[c]{@{}l@{}}polatgorkem \\(METU\_DLCV)\end{tabular} & \begin{tabular}[c]{@{}l@{}}Faster RCNN +\\CascadeRCNN +\\Retinanet\end{tabular}   & \begin{tabular}[c]{@{}l@{}}Resize\\Normalise\end{tabular}             & Ensemble                                                             & Accuracy++                                                   & \begin{tabular}[c]{@{}l@{}}ResNet50, \\ResNet101\end{tabular}             & Yes (R, F)$^\dag$                                                                         & COCO                                                   & RTX 2080                                                            & 0.76                                                     &  \href{http://github.com/GorkemP/Endoscopic-Artefact-Detection}{\nolinkurl{GorkemP/EAD}}     \\ 
\hline
\begin{tabular}[c]{@{}l@{}}qzheng5 \\(CVML)\end{tabular}           & Faster RCNN                                                                       & \begin{tabular}[c]{@{}l@{}}Resize\\Normalise\end{tabular}             & Context                                                           & Accuracy+                                                    & ResNet101                                                                 & Yes (R, T, LD)$^\dag$                                                                     & COCO                                                   & GTX1060                                                            & 0.20                                                     &  \href{http://github.com/ybguo1/CVML_EAD2020}{\nolinkurl{CVML/EAD2020}}                      \\ 
\hline
xiaohong1                                                          & \begin{tabular}[c]{@{}l@{}} YOLACT + \\NMS-within-class\end{tabular} & None                                                                  & \begin{tabular}[c]{@{}l@{}}Context \end{tabular}           & \begin{tabular}[c]{@{}l@{}}Accuracy+\\, speed+\end{tabular}  & ResNet101                                                                 & None                                                                               & ImageNet                                               & Tesla K80                                                          & 0.14                                                     &  
\href{http://github.com/dbolya/yolact}{\nolinkurl{yolact}}\\ 
\hline
mathew666                                                          & \begin{tabular}[c]{@{}l@{}}Faster RCNN +\\~NMS\end{tabular}                       & None                                                                  & Context                                                           & Accuracy+                                                    & ResNet101                                                                 & Yes                                                                                & NA                                                     & RTX 2080                                                           & NA                                                       & NA                                          \\ 
\hline
VinBDI                                                             & EfficientDet D0                                                                   & \begin{tabular}[c]{@{}l@{}}Resize \\(512x512)\end{tabular}            &\begin{tabular}[c]{@{}l@{}}Multiscale\\scalable \end{tabular}                                                                & Speed++                                                      & EfficientNet B0                                                           & \begin{tabular}[c]{@{}l@{}}Yes (S, Sc, R, \\N, MU)$^\dag$\end{tabular}                    & COCO                                                   & RTX 2080TI                                                         & NA                                                       & \href{http://github.com/VinBDI-MedicalImagingTeam/endocv2020-seg}{\nolinkurl{endocv2020-seg}}    \\ 
\hline
higersky                                                           & Cascade R-CNN                                                                     & None                                                                  & Cascading                                                            & Accuracy++                                                   & ResNeXt101                                                                & Yes                                                                                & NA                                                     & GTX1080 Ti                                                         & NA                                                       & NA                                          \\ 
\hline
StarStarG                                                          & Cascade R-CNN                                                                     & \begin{tabular}[c]{@{}l@{}}Resize\\Normalise\end{tabular}             & Cascading                                                            & Accuracy++                                                   & ResNeXt101                                                                & Yes (F, S)$^\dag$                                                                         & NA                                                     & RTX 2080                                                           & NA                                                       & NA                                          \\ 
\hline
anand\_subu                                                        & RetinaNet                                                                         & \begin{tabular}[c]{@{}l@{}}Resize\\Normalise\end{tabular}             & Context                                                              & \begin{tabular}[c]{@{}l@{}}Accuracy+\\, speed+\end{tabular}  & \begin{tabular}[c]{@{}l@{}}ResNet101 \end{tabular} & \begin{tabular}[c]{@{}l@{}}Yes (R, Sh, F, C, \\B, St, H)$^\dag$\end{tabular}              & ImageNet                                               & \begin{tabular}[c]{@{}l@{}}GTX1050Ti \end{tabular} & \begin{tabular}[c]{@{}l@{}}0.36 \end{tabular}  &                                                   \href{http://github.com/anand-subu/EAD2020-Challenge-Code}{\nolinkurl{anand-subu/EAD2020}}
\\ 
\hline
arnavchavan04                                                      & \begin{tabular}[c]{@{}l@{}}RetinaNet + \\FasterRCNN \\(FPN + DC5) \end{tabular}   & \begin{tabular}[c]{@{}l@{}}Resize \\(512x512)\end{tabular}            & Ensemble                                                             & Accuracy++                                                   & \begin{tabular}[c]{@{}l@{}}ResNet50; \\ResNeXt101\end{tabular}            & Yes (F, C, R)$^\dag$                                                                      & ImageNet                                               & Tesla T4                                                          & NA                                                       & \href{http://github.com/ubamba98/EAD2020}{\nolinkurl{ubamba98/EAD2020}}                            \\ 
\hline
MXY                                                                & \begin{tabular}[c]{@{}l@{}}Cascase RCNN + \\FPN\end{tabular}                      & \begin{tabular}[c]{@{}l@{}}Resize\\Normalise\end{tabular}             & Cascading                                                            & Accuracy+                                                    & ResNet101                                                                 & Yes (F)$^\dag$                                                                            & ImageNet                                               & RTX 2080 Ti                                                         & 0.80                                                     & \href{http://github.com/Carboxy/EAD2020}{\nolinkurl{Carboxy/EAD2020}}                             \\ 
\hline
mimykgcp                                                           & \begin{tabular}[c]{@{}l@{}}Faster RCNN + \\+ RetinaNet\end{tabular}               & \begin{tabular}[c]{@{}l@{}}Resize\\Normalise\end{tabular}             & Ensemble                                                             & \begin{tabular}[c]{@{}l@{}}Accuracy+\\, speed+\end{tabular}  & ResNeXt101                                                                & Yes (RA)$^\dag$                                                                           & COCO                                                   & GTX 1080Ti                                                         & 0.58                                                     & NA                                          \\ 
\hline
\begin{tabular}[c]{@{}l@{}}DuyHUYNH \\(LRDE)\end{tabular}          & YOLOv3                                                                            & Normalise                                                             & Multiscale                                                           & \begin{tabular}[c]{@{}l@{}}Accuracy+\\, speed++\end{tabular} & Darknet53                                                                 & Yes (RA)$^\dag$                                                                           & COCO                                                   & GTX1080 Ti                                                         & 0.07                                                     & \href{https://gitlab.lrde.epita.fr/dhuynh/endocv2020}{\nolinkurl{dhuynh/endocv2020}}                          \\ 
\hline
\multicolumn{11}{l}{\begin{tabular}[c]{@{}l@{}} \textbf{}\\\textbf{Segmentation } \end{tabular}}                                                                                                            \\ 
\hline
\begin{tabular}[c]{@{}l@{}}qzheng5 \\(CVML)\end{tabular}           & DeepLabv3+                                                                        & \begin{tabular}[c]{@{}l@{}}Resize \\(513x513)\\Normalise\end{tabular} & \begin{tabular}[c]{@{}l@{}}Encoder-decoder,~\\mutiscale\end{tabular} & Accuracy++                                                   & SE-ResNeXt50                                                              & (R, T, LD + TTA)$^\dag$~                                                                    & ImageNet                                               & GTX1080Ti~                                                         & \begin{tabular}[c]{@{}l@{}}0.50; \\5 (+TTA)\end{tabular} & \href{http://github.com/ybguo1/CVML_EAD2020}{\nolinkurl{CVML/EAD2020}}                        \\ 
\hline
mouradai\_ox                                                       & Pyramid dilated module                                                            & \begin{tabular}[c]{@{}l@{}}Resize \\(512x512)\\Normalise\end{tabular} & Multiscale                                                           & \begin{tabular}[c]{@{}l@{}}Accuracy+~\\, speed+\end{tabular} & ResNet50                                                                  & Yes (T, R, LD)$^\dag$                                                                           & ImageNet                                               & Colab                                                              & 0.37                                                     & NA                                          \\ 
\hline
arnavchavan04                                                      & \begin{tabular}[c]{@{}l@{}}FPN + \\EfficientNet~\end{tabular}                     & \begin{tabular}[c]{@{}l@{}}Resize \\(512x512)\end{tabular}            & Ensemble                                                             & Accuracy+                                                    & EfficientNet                                                              & Yes (F, C, R)$^\dag$                                                                      & ImageNet                                               & Tesla T4                                                          & NA                                                       & \href{http://github.com/ubamba98/EAD2020}{\nolinkurl{ubamba98/EAD2020}}                           \\ 
\hline
VinBDI                                                             & \begin{tabular}[c]{@{}l@{}}U-Net + \\BiFPN~\end{tabular}                          & \begin{tabular}[c]{@{}l@{}}Resize \\(512x512)\end{tabular}            & \begin{tabular}[c]{@{}l@{}}Ensemble,\\Endcoder-decoder\end{tabular}  & \begin{tabular}[c]{@{}l@{}}Accuracy++\\, speed+\end{tabular} & \begin{tabular}[c]{@{}l@{}}EfficientNet B4;~\\ResNet50\end{tabular}       & Yes (S, Sc, R, F)$^\dag$                                                                   & \begin{tabular}[c]{@{}l@{}}COCO\\ImageNet\end{tabular} & RTX 2080TI                                                         & NA                                                       & \href{http://github.com/VinBDI-MedicalImagingTeam/endocv2020-seg}{\nolinkurl{endocv2020-seg}}   \\ 
\hline
higersky                                                           & DeepLabv3+                                                                        & None                                                                  & \begin{tabular}[c]{@{}l@{}}Encoder-decoder,\\mutiscale\end{tabular}  & Accuracy+                                                    & ResNet101                                                                 & Yes (F;S;Sc;Bl)$^\dag$                                                                    & ImageNet                                               & GTX1080 Ti                                                         & NA                                                       & NA                                          \\ 
\hline
anand\_subu                                                        & U-Net                                                                             & \begin{tabular}[c]{@{}l@{}}Resize \\(512x512)\end{tabular}                                                                       & Encoder-decoder                                                      & Accuracy+                                                    & ResNet50                                                                  & \begin{tabular}[c]{@{}l@{}}Yes (S, F, R, N, \\Cr, Bl, H, St, \\C, Sp)$^\dag$\end{tabular} & ImageNet                                               & \begin{tabular}[c]{@{}l@{}}GTX1050Ti\end{tabular}     &  0.17                                                  &  \href{http://github.com/anand-subu/EAD2020-Challenge-Code}{\nolinkurl{anand-subu/EAD2020}}          \\ 
\hline
\begin{tabular}[c]{@{}l@{}}DuyHUYNH \\(LRDE)\end{tabular}          & U-Net++                                                                           & Normalise                                                             & Encoder-decoder                                                      & \begin{tabular}[c]{@{}l@{}}Accuracy+, \\speed+\end{tabular}                                              & EfficientNet B1                                                           & \begin{tabular}[c]{@{}l@{}}Yes (R, S, F, \\Sc,~LD, TTA)$^\dag$\end{tabular}               & ImageNet                                               & GTX1080 Ti                                                         & 0.97                                                     & \href{https://gitlab.lrde.epita.fr/dhuynh/endocv2020}{\nolinkurl{dhuynh/endocv2020}}                        \\ 
\hline
mimykgcp                                                           & U-Net                                                                             & \begin{tabular}[c]{@{}l@{}}Resize\\Normalise\end{tabular}             & Encoder-decoder                                                      & \begin{tabular}[c]{@{}l@{}}Accuracy+, \\speed+\end{tabular}  & ResNeXt50                                                                 & Yes (RA)$^\dag$                                                                           & ImageNet                                               & RTX 2070                                                           & 0.25                                                     & NA                                          \\ 
\hline
\bottomrule
\multicolumn{10}{l}{$^\dag$ B: brightness, C: contrast, F: Flip, H: hue, LD: Local deformation, N: noise, R: Rotation, RA: RandAugment, S: Shift, Sc: scaling Sh: shear,} \\
\multicolumn{10}{l} {St: saturation, Mu: mixup, T: Translation, TTA: test-time augmentation} \hspace{.1cm}\\
	   %$^\diamond$https://github.com/ \hspace{.1cm} $^\ast$ https://gitlab.lrde.epita.fr/}\\
\hline
\end{tabular}
}
\end{adjustbox}
\end{table*}
\begin{table*}
\caption{Endoscopy disease detection and segmentation (EDD2020) method summary for top 7 teams (out-of 14 submission).\label{tab:EDD2020_method_summary}}
\centering
\begin{adjustbox}{width=1\textwidth}
\small{
\begin{tabular}{l|l|l|l|l|l|l|l|l|l|l} 
\hline\hline
\multicolumn{1}{l|}{\textbf{Team EDD2020}}                                                & \multicolumn{1}{l|}{\textbf{Algorithm}}                                                            & \multicolumn{1}{l|}{\textbf{Preprocessing}}                                               & \multicolumn{1}{l|}{\textbf{Nature}}                                                     & \multicolumn{1}{l|}{\textbf{Basis-of-choice}}                                     & \multicolumn{1}{l|}{\textbf{Backbone}}                                                 & \multicolumn{1}{l|}{\textbf{Data aug.}}                                                   & \multicolumn{1}{l|}{\textbf{Pretrained}}                                                   & \multicolumn{2}{c|}{\textbf{Computation}} & \multicolumn{1}{c}{\textbf{code}}                                                                                \\ 
\hline\hline
\multicolumn{8}{l|}{\begin{tabular}[c]{@{}l@{}}\textbf{}\\\textbf{Detection~} \end{tabular}}                                                     & \multicolumn{1}{l|}{\textbf{GPU}}                                 & \multicolumn{1}{l|}{\textbf{Test time}}                 &                                             
 \\ 
\hline
Adrian                                                               & \begin{tabular}[c]{@{}l@{}}YOLOv3+\\Faster R-CNN\end{tabular}                  & Resize                                                               & Ensemble                                                           & \begin{tabular}[c]{@{}l@{}}Accuracy+\\, speed+\end{tabular}  & \begin{tabular}[c]{@{}l@{}}Darnet53\\ResNet101\end{tabular}       & Yes (F, D)$^\dag$                                                           & \begin{tabular}[c]{@{}l@{}}COCO\\public polyp \\dataset\end{tabular}   & Tesla P100         & 0.41                       & \href{https://github.com/Adrian398/Endoscopic\_Disease\_Detection}{\nolinkurl{Adrian398/EDD}}                                  \\ 
\hline
shahadate                                                            & Mask R-CNN                                                                     & \begin{tabular}[c]{@{}l@{}}Resize\\Normalise\end{tabular}            & Multiscale                                                          & \begin{tabular}[c]{@{}l@{}}Accuracy\\, speed+\end{tabular}   & ResNet101                                                         & \begin{tabular}[c]{@{}l@{}}Yes (Sc, R, F, \\Cr, S, N)$^\dag$\end{tabular}    & COCO                                                                   & RTX2060      & NA                         & \href{https://github.com/Shahadate-Rezvy/EDD-Mask-rcnn}{\nolinkurl{EDD-Mask-rcnn}}                                             \\ 
\hline
VinBDI                                                               & EfficientDet D0                                                                & \begin{tabular}[c]{@{}l@{}}Resize \\(512x512)\end{tabular}           & Ensemble                                                            & Speed++                                                      & EfficientNet B0                                                   & \begin{tabular}[c]{@{}l@{}}Yes (S, Sc, \\R, N, MU)$^\dag$\end{tabular}      & COCO                                                                   & RTX 2080TI    & NA                         & \href{https://github.com/VinBDI-MedicalImagingTeam/endocv2020-seg}{\nolinkurl{endocv2020-seg}}                                 \\ 
\hline
YH\_Choi                                                             & CenterNet                                                                      & NA                                                                   & Context                                                             & Accuracy++                                                   & ResNet50                                                          & \begin{tabular}[c]{@{}l@{}}Yes(Du, R, \\F, C, B)$^\dag$\end{tabular}        & \begin{tabular}[c]{@{}l@{}}PASCAL \\VOC2012\end{tabular}               & RTX 2080      & 2                          & NA                                                                                           \\ 
\hline
\begin{tabular}[c]{@{}l@{}}DuyHUYNH \\(LRDE)\end{tabular}            & U-Net++                                                                        & Normalise                                                            & Encoder-decoder                                                     & Speed                                                        & EfficientNet B1                                                   & \begin{tabular}[c]{@{}l@{}}Yes (R, S, F, \\Sc, LD, TTA)$^\dag$\end{tabular} & ImageNet                                                               & GTX1080 Ti   & 1.53                       & \href{https://gitlab.lrde.epita.fr/dhuynh/endocv2020}{\nolinkurl{dhuynh/endocv2020}  }                                             \\ 
\hline
\begin{tabular}[c]{@{}l@{}}mimykgcp\\(vishnusai)\end{tabular}        & \begin{tabular}[c]{@{}l@{}}Faster RCNN +\\RetinaNet\end{tabular}               & \begin{tabular}[c]{@{}l@{}}Resize \\(256x256) normalise\end{tabular} & Ensemble                                                            & \begin{tabular}[c]{@{}l@{}}Accuracy+\\, speed+\end{tabular}  & ResNeXt101                                                        & Yes (RA)$^\dag$                                                             & COCO                                                                   & GTX1080Ti    & 0.58                       & NA                                                                                           \\ 
\hline
\multicolumn{11}{l}{\begin{tabular}[c]{@{}l@{}} \textbf{}\\\textbf{Segmentation } \end{tabular}}                                                                                                            \\ 
\hline
Adrian                                                               & \begin{tabular}[c]{@{}l@{}}YOLOv3 +\\Faster R-CNN +\\Cascade RCNN\end{tabular} & Resize                                                               & \begin{tabular}[c]{@{}l@{}}Ensemble\end{tabular}        & Accuracy++                                                   & \begin{tabular}[c]{@{}l@{}}Darnet53\\ResNet101\end{tabular}       & Yes (F, D)$^\dag$                                                           & \begin{tabular}[c]{@{}l@{}}COCO \\public polyp \\dataset\end{tabular} & Tesla P100         &                            & \href{https://github.com/Adrian398/Endoscopic\_Disease\_Detection}{\nolinkurl{Adrian398/EDD2020}}                                  \\ 
\hline
shahadate                                                            & MaskRCNN                                                                       & \begin{tabular}[c]{@{}l@{}}Resize\\Normalise\end{tabular}            & Multiscale                                                          & \begin{tabular}[c]{@{}l@{}}Accuracy\\, speed+\end{tabular}   & ResNet101                                                         & \begin{tabular}[c]{@{}l@{}}Yes (Sc, R, F, \\Cr, S, N)$^\dag$\end{tabular}   & COCO                                                                   & RTX2060      &                            & \href{https://github.com/Shahadate-Rezvy/EDD-Mask-rcnn}{\nolinkurl{EDD-Mask-rcnn}}                                             \\ 
\hline
VinBDI                                                               & \begin{tabular}[c]{@{}l@{}}U-Net + \\BiFPN~\end{tabular}                       & Resized (512x512)                                                    & \begin{tabular}[c]{@{}l@{}}Ensemble\\Endcoder-decoder\end{tabular} & \begin{tabular}[c]{@{}l@{}}Accuracy++\\, speed+\end{tabular} & \begin{tabular}[c]{@{}l@{}}EfficientNet B4\\ResNet50\end{tabular} & \begin{tabular}[c]{@{}l@{}}Yes (S, Sc, \\R, F)$^\dag$\end{tabular}          & \begin{tabular}[c]{@{}l@{}}COCO\\ImageNet\end{tabular}                 & RTX 2080 Ti   & NA                         & \href{https://github.com/VinBDI-MedicalImagingTeam/endocv2020-seg}{\nolinkurl{endocv2020-seg}}                                  \\ 
\hline
YH\_Choi                                                             & U-Net & NA       & Encoder-decoder              &              Accuracy+          & ResNet50                    & \begin{tabular}[c]{@{}l@{}}Yes(Du, R, F, \\C, B)$^\dag$\end{tabular}        & \begin{tabular}[c]{@{}l@{}}PASCAL \\VOC2012\end{tabular}               & RTX 2080      & 7  & NA    \\ 
\hline
\begin{tabular}[c]{@{}l@{}}DuyHUYNH \\(LRDE)\end{tabular}            & U-Net++   & Normalise & Encoder-decoder               & \begin{tabular}[c]{@{}l@{}}Accuracy+\\, speed+\end{tabular}  & EfficientNet B1  & \begin{tabular}[c]{@{}l@{}}Yes (R, S, F, \\Sc, LD, TTA)$^\dag$\end{tabular} & ImageNet  & GTX1080 Ti   & 1.53  & \href{https://gitlab.lrde.epita.fr/dhuynh/endocv2020}{\nolinkurl{endocv2020}}                                               \\
\hline
\begin{tabular}[c]{@{}l@{}}drvelmuruganb\end{tabular} & SUMNet   & NA & Encoder-decoder   & \begin{tabular}[c]{@{}l@{}}Accuracy+\\, speed++\end{tabular}                                                              & VGG11                                                             & \begin{tabular}[c]{@{}l@{}}Yes(R, A, Sc, \\P, and Cr)$^\dag$ \end{tabular}   & ImageNet                                                               & GTX1080 Ti   & 0.16                      & \href{https://github.com/drvelmuruganb/EDD2020-Endoscopy-disease-detection-grand-challenge-2020}{\nolinkurl{drvelmuruganb/EDD2020}}  \\ 
\hline
\begin{tabular}[c]{@{}l@{}}mimykgcp \end{tabular}          & U-Net & \begin{tabular}[c]{@{}l@{}}Resize\\Normalise\end{tabular}            & Encoder-decoder  & Accuracy+ & ResNeXt50 & Yes (RA)$^\dag$  & ImageNet  & RTX2070      & 1.25                       & NA                                                                                           \\ 
\hline
\bottomrule
\multicolumn{11}{l}{$^\dag$ A: affine, B: brightness, C: contrast, Cr: cropping, D: distortion, Du: duplication, F: flip, H: hue, LD: local deformation, Mu: mixup, N: noise, P: perspective transformation, R: rotation, RA: RandAugment library,} \\
\multicolumn{11}{l} {S: shift, Sc: scaling, Sh: shear, St: saturation, T: translation, TTA: test-time augmentation} \\
\hline
\end{tabular}
}
\end{adjustbox}
\end{table*}
%\end{landscape}
%
%
In this Section, we present summary of top participating teams for both EAD2020 and EDD2020 sub-challenges. Each of these teams has participated in either detection task or segmentation task or both.
\subsection{EAD2020 Participating teams}
%
% Also prrofread the methods description please
\begin{itemize}
     \item \textbf{Team \textit{polatgorkem}}~\cite{metu}
    The team used an ensemble of three object detectors: Faster R-CNN (ResNet50 with FPN), Cascade R-CNN (ResNet50 with FPN), RetinaNet (ResNet101 with FPN). Class-agnostic NMS operation, where the model predictions were passed through the NMS procedure together for all classes, was applied to the output of each individual model. During ensemble, only the bounding boxes for which majority of the models agree were kept. False-positive elimination was applied as a post-processing step to eliminate same-type predicted boxes located close to each other. For each class, an IoU threshold was determined. 
%When there were predicted bounding boxes with higher IoU values than the threshold, the ones having lower confidence scores were removed. 
% Horizontal flipping and 90, 180, 270 degrees rotation were applied to increase variance in the dataset.  %Models were trained on two NVIDIA RTX2080 GPUs.
    
        \item \textbf{Team \textit{CVML}}~\cite{cvml}
        CVML team's model was inspired by DeepLabV3+. The team experimented with several changes including the backbone, the global pooling, the dilated kernels and the convolution kernels with dilation rates. Moreover, the squeeze-and-excitation module is added behind the balanced ASPP module to introduce attention gating at the output of the original encoder to better utilize the information available in the computed feature maps. In addition, the original multi-class classifier is replaced with 5 binary classifiers to enable segmentation of the overlapping objects. At test time, they used some post-processing techniques such as rotation, holes filling and removal of objects from the image boundary. 

    \item \textbf{Team \textit{{mouradai\_ox}}}~\cite{OXEndonet}
The team proposed a novel neural network called OxEndoNet to tackle the segmentation challenge. The network uses the pyramid dilated module (PDM) consisting of multiple dilated convolutions stacked in parallel. For each input image, pre-trained ResNet50 (on ImageNet) was used as the backbone to extract the feature map followed by multiple PDM layers to form an end-to-end trainable network. In the final architecture, they used four PDM layers; each layer used four parallel dilated convolutions with a filter size of \(3 \times 3\) and dilation rates of 1, 2, 3, and 4. They fed the final PDM layer to a convolution layer followed by a bilinear interpolation to up-scale the feature map to the original image size.
    
\item \textbf{Team \textit{mimykgcp}}\cite{mimyk} 
   The team re-trained the ResNeXt101 backbone with the cardinality parameter set to 64. To enable detection of artefacts at different scales, an FPN was integrated into the object detectors. Data-Augmentation techniques based on RandAugment~\cite{cubuk2019randaugment} were incorporated to improve the generalization capability. For the segmentation task, a U-Net with an ImageNet pre-trained ResNext50 backbone was used.  
   
    \item \textbf{Team \textit{DuyHUYNH}}~\cite{lrde}
    For segmentation, the team exploited a model based on U-Net++ using pre-trained EfficientNet on ImageNet as the backbone. The model was trained to minimize F2-loss using the Adam optimizer. At the test-time the team used five transformations: horizontal, vertical flipping, and three rotations. For detection, the team used the bounding boxes deduced from the results of their segmentation model on the EDD dataset, while for EAD, they used YOLOv3 pre-trained on COCO.
    
%Various transformations were used such as rotation, RGB value shift, horizontal or vertical flipping, random scaling, elastic deformation, and random cropping. Inputs were scaled between 0 and 1, and then each channel was normalized using ImageNet's mean and standard deviation.  Horizontal and vertical flipping were used for data augmentation and images were scaled between 0 and 1. %YOLOv3 was trained on TITAN X (Pascal).
    
      \item \textbf{Team \textit{mathew666}}~\cite{HongyuEAD2020}%Hongyu Hu , Shanghai Jiaotong University, mathewcrespo@sjtu.edu.cn
 The team used Cascade RCNN architecture with the ResNeXt backbone in a FPN based feature extraction paradigm. Data augmentation with probability of 0.5 for horizontal flip was applied.  The team also utilised multi-scale detection to tackle with variable sized object detection.
    
  \item \textbf{Team \textit{arnavchavan04}}~\cite{paperid_20}
  % Multiplateu arnav and rishav tiwari Paperid\_20
 For the object detection task, the team used an ensemble of three models: Faster R-CNN (ResNext101 + FPN), RetinaNet (ResNet101 + FPN) and Faster R-CNN (ResNext101 + DC5). For the segmentation task, an ensemble of multiple depth EfficientNet models with FPN trained on multiple optimization plateaus (DSC, BCE, IoU) was designed. Data augmentation techniques like horizontal and vertical flip, cutout (random holes), random contrast, gamma, brightness, rotation along with CutMix~\cite{yun2019cutmix} strategy for the segmentation task were incorporated to improve generalization capability.
 
      \item \textbf{Team \textit{anand\_subu}}~\cite{askm}
 %Anand subramanian
The team used RetinaNet with ResNet101 backbone. For the segmentation task, the team used an ensemble network with U-Net with a ResNet50 backbone and DeepLabV3. However, the team reported U-Net with ResNet101 as their best architecture of choice. All the backbones were pre-trained on the ImageNet. Real-time augmentation techniques like rotation, shear, random-image-flip, image contrast, brightness, saturation, and hue variations were incorporated while training to improve the generalization capability of the network. 

 \item \textbf{Team \textit{higersky}}~\cite{paperid_26}
 % Haijian chen higersky
 The team implemented Hyper Task Cascade and Cascade R-CNN with ResNeXt101 backbone as a feature extractor and FPN module for multi-scale feature representation for the object detection task. They applied Soft-NMS~\cite{bodla2017soft} to avoid mistakenly discarded bounding-boxes. For the semantic segmentation task, the team incorporated DeepLabV3+ with ResNet101 backbone and trained with BCE and DICE losses. The backbones for both tasks were pre-trained on ImageNet.
 
 \item \textbf{Team \textit{MXY}}~\cite{paperid_27}
 % Zhimiao yu and Yuanfan Guo 
 The team used a Cascade R-CNN with an ImageNet pre-trained ResNet101 backbone and a FPN module. Post-detection, soft-NMS was added to remove false predictions. The dataset was augmented by random resizing technique to improve the final output scores. The team used more weight for the losses of specularity, artefact, and bubbles classes to overcome classification difficulties between those classes.
 
 %TODO arXiv add?
\item \textbf{Team \textit{StarStarG}}
The team used Cascade-RCNN as network architecture and adopted COCO2017 pre-trained ResNeXt as backbone with FPN and multi-stage RCNN framework. The authors also integrated Deformable Convolutional Networks in backbone to improve the model performance. 

\item \textbf{Tesam \textit{xiaohong1}}~\cite{xiaohong1}
   The team built their detection and segmentation method upon Yolact-based instance segmentation system. Yolact~\cite{yolact-iccv2019} adds a segmentation component to the RetinaNet to ensure the tasks of detection, classification and delineation which are performed simultaneously. The network uses ResNet101 as an imageNet pretrained backbone.
 
% \item \textbf{Team 11: Paperid\_31\cite{paperid_31}}
% % Hoang Manh Hung, Phan Tan Dac Thinh
% For the object detection task, the authors devised an ensemble of Cascade R-CNN (COCO pre-trained ResNext101 backbone + DCN) models trained at 5 different folds of the entire competition dataset. NMS was employed post-detection to remove redundant detections. For the segmentation task, U-Net was chosen along with a SE-ResNext 50 backbone taking to factors like performance and speed of training. The authors stated that augmentation techniques were not used.
 
 % short papers: we will include 1 author from this 
 % selection criteria is based on their result ranking
% \item \textbf{Team 13: xiaohong1~\cite{xiaohong1}}
%  %xiaohong1: Middlesex Univ. London
% The authors used Yolact-based architecture. 
\end{itemize}

\subsection{EDD2020 Participating teams}
\begin{itemize}
 \item \textbf{Team \textit{Adrian}}~\cite{krenzer}  The team compared two different models: YOLOv3 with darknet-53 backbone and Faster R-CNN with ResNet-101 backbone. For post-processing, both algorithms in the final architecture were combined. For the second task, the team leveraged the state-of-the-art Cascade Mask R-CNN with ResNeXt-151 as a backbone. The team trained YOLOv3 using categorical cross-entropy for classification and default localization loss, while for Cascade Mask-RCNN, they used binary cross entropy for classification and mask, and L1 smooth for boundary box regression. %As a post-processing step, the team used the knowledge from the endoscopy domain to further refine their models and compare the resulting performances. 
%All models were trained on a Tesla P100 GPU. 

 \item \textbf{Team \textit{Shahadate}}~\cite{rezvy}
The team implemented a modified benchmark Mask R-CNN infrastructure model on the EDD2020 dataset. They used COCO trained weights and biases with the ResNet101 backbone as an initial feature extractor. The network head of the backbone model was replaced with new untrained layers that consisted of a fully-connected classifier with five classes and an additional background class. Non-maximum suppression was used to reduce overlapped detection. Finally, the team merged multiple bounding boxes for the same class label as one bounding box to match with the mask annotation. 
%Common data augmentation techniques were used such as instance cropping, random rotation and crop, random shifting and scaling, horizontal and vertical flip with Gaussian noise. 

 \item \textbf{Team \textit{VinBDI}}~\cite{vinbdi}
 For the object detection task, the team designed an ensemble of six EfficientDet models (with BiFPN modules) trained on six different EfficientNet backbones. A total of eleven augmentation techniques were incorporated to increase the output prediction scores of the model. For the segmentation task, an ensemble of U-Net and EfficientNet-B4 and BiFPN with the ResNet50 backbone was devised. The same team also participated in the EAD2020 sub-challenge.

 \item \textbf{Team \textit{YH\_Choi}}~\cite{yhchoi}
 The team implemented a CenterNet-based model with the PASCAL VOC pretrained ResNet50 backbone for the object detection task. A similar backbone with U-Net was devised for the segmentation task. The dataset was randomly duplicated to tackle class-imbalance. To improve generalization performance, each image was augmented 86 times by randomly choosing augmentation techniques from the pool of rotation, flipping, contrast enhancement and brightness adjustment.

\item \textbf{Team \textit{drvelmuruganb}}~\cite{VelmuruganEDD2020}
For the segmentation of disease classes the team used an encoder-decoder based SUMNet architecture with the ImageNet pretrained VGG11 backbone. The authors also applied several augmentation strategies including variable brightness and HSV values, multiple crops and geometric transformations such as rotation, affine, scaling and projective were also applied to improve the accuracy. 
\end{itemize}
\begin{table}[t!]
\small{
\caption{\textbf{EAD2020 results for the detection task on the single frame dataset.} mAP at IoU thresholds 25\%, 50\% and 75\% are provided along with overall mAP and overall IoU computations. Overall scores are computed at 11 IoU thresholds and averaged. Weighted detection score $score_d$ is computed between overall mAP and IoU scores only. Three best scores for each metric criteria are in bold.\label{tab:ead2020_detection_singleframe}}
\resizebox{\columnwidth}{!}{
\begin{tabular}{l|c|c|c|c|c|c|c}
\hline
\multicolumn{1}{c|}{\textbf{\begin{tabular}[c]{@{}c@{}}Team \\ names\end{tabular}}} & \textbf{mAP$_{25}$} & \textbf{mAP$_{50}$} & \textbf{mAP$_{75}$} & \textbf{\begin{tabular}[c]{@{}c@{}}overall \\mAP$_{d}$\end{tabular}} & \textbf{\begin{tabular}[c]{@{}c@{}}overall\\ mIoU$_{d}$\end{tabular}} & \textbf{mAP$_\delta$} & \textbf{score$_d \pm \delta$ }  \\ \hline \hline%
\begin{tabular}[c]{@{}l@{}}polatgorkem\end{tabular}  & 26.886 & 17.883 & 5.608 & 17.486 &  \textbf{36.579} & 7.124 &  \textbf{25.123 $\pm$ 7.124 }\\ \hline
\begin{tabular}[c]{@{}l@{}}qzheng5 \end{tabular} & 33.134 & 20.084 & 5.570 & 19.720 & 27.185 & 8.820 & 22.706 $\pm$ 8.820 \\ \hline
xiahong1 & 30.627 & 19.384 & 4.935 & 18.512 & 26.388 & 8.428 & 21.663 $\pm$ 8.428 \\ \hline
mathew666 & 20.360 & 19.440 & 7.783 & 18.091 &  \textbf{32.692} & 5.617 &  \textbf{23.931 $\pm$ 5.617} \\ \hline
VinBDI & 38.429 & 25.426 & 7.053 & 24.069 & 12.644 &  \textbf{10.291} & 19.499 $\pm$ 10.291 \\ \hline
higersky & 36.920 & 25.770 &  \textbf{9.452} & 24.771 & 17.298 & 8.707 & 21.781 $\pm$ 8.707 \\ \hline
StarStarG & \textbf{41.800} &  \textbf{29.984} &  \textbf{10.733} &  \textbf{28.380} & 16.250 &  \textbf{10.042 }& 23.528 $\pm$ 10.042 \\ \hline
anand\_subu & 29.755 & 19.893 & 5.271 & 18.886 & 24.029 & 7.619 & 20.943 $\pm$ 7.619 \\ \hline
arnavchavan04 &  \textbf{38.752} &  \textbf{27.247} &  \textbf{9.858} &  \textbf{26.021} & 21.165 & 9.342 &  \textbf{24.079 $\pm$ 9.342} \\ \hline
MXY & 25.373 & 18.967 & 7.171 & 17.82 &  \textbf{28.056} & 5.754 & 21.914 $\pm$ 5.754 \\ \hline
mimykgcp &  \textbf{39.897} &  \textbf{26.296} & 6.839 &  \textbf{25.082} & 10.209 &  \textbf{10.765} & 19.133 $\pm$ 10.765 \\ \hline
DuyHUYNH & 20.512 & 12.234 & 2.978 & 11.894 & 27.063 & 5.671 & 17.962 $\pm$ 5.671 \\ \hline
\hline
\multicolumn{8}{l}{\textbf{baselines} }\\ 
\hline \hline
%Yolov3 & 22.656 & 14.043 & 2.804 & 13.249 & 24.882 & 6.524 & 17.902 $\pm$ 6.524 \\ \hline
YOLOv3	& 22.798  &	13.736 &	2.804 &	13.249  &	24.883	& 6.525 &	17.903 $\pm$ 6.525	\\ \hline
\begin{tabular}[c]{@{}l@{}}RetinaNet\\(ResNet101)\end{tabular} &	15.270 &	8.927&	2.061&	8.754&	23.202&	4.275&	14.533 $\pm$ 4.275	
\\ \hline
% generalisation
% Yolov3-spp & 24.549 & 16.278 & 4.468 & 15.456 & 19.821 & 6.628 & 17.202 $\pm$ 6.628 \\ \hline
%
\end{tabular}
}
}
\end{table}
%%%%%
%%%%
%%%%%
\begin{table}[!t]
\caption{\textbf{EAD2020 results for the sequence dataset.} mAP at IoU thresholds 25\%, 50\% and 75\% are provided along with overall mAP and overall IoU computations. Overall scores are averaged with 11 IoU thresholds. Weighted detection score $score_d$ is computed between overall mAP and IoU scores only. Three best scores for each metric criteria are in bold.\label{tab:ead2020_detection_sequence}}
\centering
\small{
\resizebox{\columnwidth}{!}{
\begin{tabular}{l|c|c|c|c|c|c|c}
\hline
\multicolumn{1}{c|}{\textbf{\begin{tabular}[c]{@{}c@{}}Team \\ names\end{tabular}}} & \textbf{mAP$_{25}$} & \textbf{mAP$_{50}$} & \textbf{mAP$_{75}$} & \textbf{\begin{tabular}[c]{@{}c@{}}overall \\mAP$_{seq}$\end{tabular}} & \textbf{\begin{tabular}[c]{@{}c@{}}overall\\ mIoU$_{seq}$\end{tabular}} & \textbf{mAP$_\delta$} & \textbf{score$_d \pm \delta$ } \\ \hline
\begin{tabular}[c]{@{}l@{}}polatgorkem\end{tabular}   & 38.464 & 24.803 & 4.138 & 23.137 &  \textbf{29.117} & 10.326 &  \textbf{25.529 $\pm$ 10.326} \\ \hline
\begin{tabular}[c]{@{}l@{}}qzheng5 \end{tabular}  &  \textbf{48.210} & 25.717 & 3.997 & 25.665 & 20.949 &  \textbf{14.222} &  \textbf{23.779 $\pm$ 14.222} \\ \hline
xiahong1 & 46.087 & 25.813 & 2.684 & 25.136 & 18.398 & \textbf{ 15.128} & 22.441 $\pm$ 15.128 \\ \hline
mathew666 & 31.599 & 21.878 & 3.053 & 19.623 & 20.858 & 9.718 & 20.117 $\pm$ 9.718 \\ \hline
VinBDI & 45.295 &  \textbf{26.723} & 4.396 & 25.285 &  \textbf{23.426} &  \textbf{13.972 }&  \textbf{24.542 $\pm$ 13.972} \\ \hline
higersky &  \textbf{47.716} &  \textbf{29.841} &  \textbf{4.473} &  \textbf{28.334} & 12.865 & 14.579 & 22.147 $\pm$ 14.579 \\ \hline
StarStarG & \textbf{ 46.965} &  \textbf{30.202} & \textbf{ 5.432} &  \textbf{28.107} & 8.371 & 13.367 & 20.213 $\pm$ 13.367 \\ \hline
anand\_subu & 38.352 & 25.535 & 3.843 & 23.014 & 20.703 & 10.859 & 22.089 $\pm$ 10.859 \\ \hline
arnavchavan04 & 34.511 & 21.524 &  \textbf{4.886} & 20.700 & 11.827 & 9.839 & 17.151 $\pm$ 9.839 \\ \hline
MXY & 31.391 & 19.838 & 3.620 & 18.601 &  \textbf{21.504} & 8.688 & 19.762 $\pm$ 8.688 \\ \hline
mimykgcp & 44.972 & 26.780 & 4.400 &  \textbf{25.937} & 6.892 & 13.697 & 18.319 $\pm$ 13.697 \\ \hline
DuyHUYNH & 28.632 & 15.524 & 0.815 & 15.468 & 16.968 & 9.381 & 16.068 $\pm$ 9.381 \\ \hline
\hline
\multicolumn{8}{l}{\textbf{baselines} }                                                                                                                                                                \\ 
\hline \hline
YOLOv3 &32.199 &	18.473	& 1.137	 & 17.176 & 	16.351 &	10.596 & 16.846 $\pm$ 10.596\\ \hline
\begin{tabular}[c]{@{}l@{}}RetinaNet\\(ResNet101)\end{tabular}   & 17.646 & 	6.447 &	0.767 & 	8.079 &	10.000 & 5.151   & 9.252 $\pm$ 5.151\\ \hline
\end{tabular}
}
}
\end{table}
\section{Results}{\label{Sec:results}}
For the EAD2020 sub-challenge, we present the results of 12 participating teams for multi-class artefact detection task and 8 teams for segmentation task. Similarly, for EDD2020 sub-challenge, we have included top 6 teams for detection and 7 teams for segmentation of multi-class diseases. In this section we present the quantitative and qualitative results for each team based on the evaluation metrics discussed in Section~\ref{Sec:evaluation_metrics}. For the EAD2020 sub-challenge, 3 different test dataset were released: 1) single-frame data for detection and segmentation, 2) sequence dataset for detection only and 3) out-of-sample data for generalization task only. For the detection task, the average of the aggregated sum of the detection scores for the single frame data and the sequence data were considered for final scoring. While, for the EDD2020 challenge only single frame detection and segmentation data were released. Below we present the result for each sub-challenges separately. 
\subsection{Quantitative results}
\subsubsection{EAD2020 sub-challenge}
In this section, the results of the participant teams in the EAD2020 challenge to detect and segment artifacts are presented. 
\paragraph{Detection task for EAD2020}

Table~\ref{tab:ead2020_detection_singleframe} and Table~\ref{tab:ead2020_detection_sequence} present the mAP values computed at different IoU thresholds (i.e., 25\%, 50\%, and 75\%), overall mAP, overall IoU, and the final score for the detection of the artefacts from single frame and sequence data, respectively. Additionally, we also provide results of baseline methods that include YOLOv3 and RetinaNet with darknet53 and ResNet101 backbones, respectively. {In Table~\ref{tab:ead2020_detection_singleframe} (i.e., single frame detection), it can be observed that the team \textit{polatgorkem} that implemented ensemble technique with Cascaded RCNN, Faster-RCNN  and RetinaNet surpassed the other teams by achieving the highest final score on the leaderboard (score$_d$, Eq.~\ref{eq:score_d}) of 25.123 $\pm$ 7.124 with the best overall mIoU of 36.579 providing a high overlap ratio between the generated bounding box with ground truth per frame. The method proposed by the team \textit{arnavchavan04} comes in the second place with score$_d$ of 24.079 $\pm$ 9.342 with 9\% more mAP than the winning team but large sacrifice in the mean IoU.  Similarly, for sequence data in Table~\ref{tab:ead2020_detection_sequence}, team \textit{ polatgorkem} maintained the first position with a final score of 25.529 $\pm$ 10.326. While the second scorer team \textit{VinBDI} suggested a method that obtained a better balanced between mAP and mIoU scores.} 

\begin{figure}[t!]
    \centering
    \includegraphics[width= 1.0\textwidth]{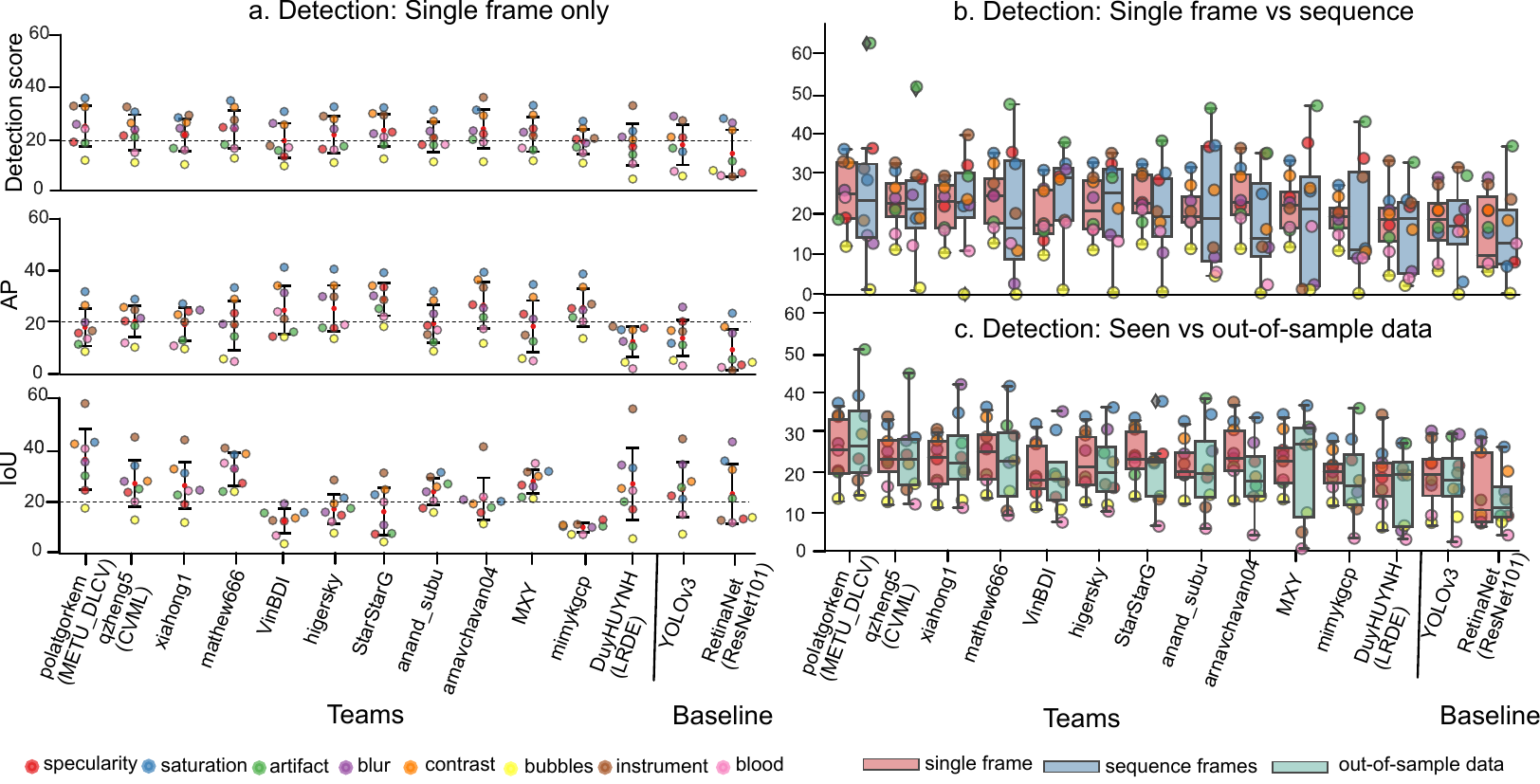}
    \caption{\textbf{Detection and out-of-sample generalization tasks for EAD2020 sub-challenge.} a) Error bars and swarm plots for the intersection over union (IoU, top), average precision (AP, middle) and challenge detection score (mAP$_d$, bottom) for each team is presented on 237 single frame test data. b-c) Comparison of mAP$_d$ w.r.t. mAP$_{seq}$ (mAP on sequence test data with 80 frames) and mAP$_{g}$ (mAP on out-of-sample data 99 frames) are provided. a-c) On the right, results from baseline detection methods: YOLOv3 and RetinaNet (with ResNet101 backbone) are also presented. Teams are arranged by decreasing overall detection ranking \textbf{R}$_{score_d}$ (see Table \ref{tab:ead2020_detection_ranking}). 
    \label{fig:ead2020_classwise_team_detection_usedscoreLeaderboard}}
\end{figure}

{Furthermore, Table~\ref{tab:ead2020_detection_ranking} shows the overall ranking for the teams in terms of Score (R$_{score_d}$), mAP (R$_{mAP}$), and generalizability performance (R$_{g}$) in addition to, mAP$_d$, mAP$_{seq}$, score$_d$, mAP$_g$ and dev$_g$. The baseline RetinaNet recorded the least deviation but also the least mAPs.  On considering the mAP$_g$ and dev$_g$ together for the final ranking of the generalization task, teams VinBDI and StarStarG secured the first place.
On observing at the class-wise performance in Figure~\ref{fig:ead2020_classwise_team_detection_usedscoreLeaderboard} (a) (i.e., single frame), it can be seen that there was a high detection score (score$_d$) and AP for larger artefact instances such as saturation and contrast. Similarly, most of the teams had a high IoU with the ground truth when detecting the instrument class. On the other hand, the detection and localization of smaller artefact instances such as  bubble and saturation showed the degraded performances by all the participating teams and by the baseline methods.}
%As shown, the best mAP$_d$ and mAP${_{g}}$ were achieved with the method proposed by team \textit{StarStarG} with values 28.380 and 25.340. For the mAP${_{seq}}$, the highest value reached 28.242 with the method proposed by the team \textit{higersky}.  Moreover, for the generalization, the method by the baseline RetinaNet, team \textit{DuyHUYNH} and YOLOv3 comes in the first, second and third places with values of 1.985, 4.807 and 4.397, respectively. On considering the mAP$_g$ and dev$_g$ together for the final ranking of the generalization task, it can be observed that teams VinBDI and StarStarG were on the first place while team higersky secured second place.

%
\begin{table}[t!]
\centering
\small{
\caption{\textbf{EAD2020 team ranking based on different metric criteria for detection and generalization task.} Overall mAPs (mAP$_d$ and mAP$_{seq}$) computed on single frame and sequence data are averaged. Final score$_d$ is then computed as the weighted value between the final IoU$_d$ and the averaged mAP. Rankings for each metric are also provided based on ascending order of the scores except for deviation score for out-of-sample data. Three best scores for each metric criteria are in bold.\label{tab:ead2020_detection_ranking}}
\resizebox{\columnwidth}{!}{
\begin{tabular}{l|c|c|c|c|c|c|c|c|c}
\hline
\begin{tabular}[c]{@{}l@{}}\textbf{Team }\\\textbf{Names}\end{tabular} & \textbf{mAP}$_d$  & \textbf{mAP}$_{seq}$ & \begin{tabular}[c]{@{}l@{}}\textbf{final }\\{\textbf{IoU}}\end{tabular}   &\begin{tabular}[c]{@{}l@{}}\textbf{final }\\{\textbf{score}$_d$}\end{tabular} & \textbf{mAP}$_g$ & \textbf{dev}$_g$ & \textbf{R}$_{score_d}$& \textbf{R}$_{mAP}$ & \textbf{R}$_{gen}$\\ 

\hline 
\hline
polatgorkem   & 17.486       & 23.137    & \textbf{32.848}        & \textbf{25.326}            & 21.008          & 9.359           & \textbf{1}               & 9                   & 6                    \\ \hline
qzheng5       & 19.720      & 24.174  & 23.751          & \textbf{22.668}            & 23.749          & 8.522           & \textbf{2}               & 6                   & 5                    \\ \hline
xiahong1      & 18.512     & 25.136   & 22.393          & \textbf{22.051}            & 24.579          & 8.169           & \textbf{3}               & 7                   & \textbf{3}           \\ \hline
mathew666     & 18.091    & 19.651  & \textbf{26.783}          & 22.035            & 16.714          & 5.674           & 4                        & 10                  & 4                    \\ \hline
VinBDI        & 24.069  & 25.282     & 18.033       & 22.018            & \textbf{24.140}          & 5.607           & 5                        & 4                   & \textbf{1}           \\ \hline
higersky      & 24.771       & \textbf{28.252}      & 15.061  & 21.931            & \textbf{24.850}          & 7.686           & 6                        & \textbf{2}          & \textbf{2}           \\ \hline
StarStarG     & \textbf{28.380}   & \textbf{28.107}  & 12.311          & 21.870            & \textbf{25.340} & 7.537           & 7                        & \textbf{1}          & \textbf{1}           \\ \hline
anand\_subu   & 18.886      & 23.004     & 22.359       & 21.510            & 20.203          & 7.896           & 8                        & 8                   & 5                    \\ \hline
arnavchavan04 & \textbf{26.021}       & 20.700  & 16.496           & 20.614            & 21.138          & 6.968           & 10                       & 5                   & \textbf{3}           \\ \hline
MXY           & 17.820      & 18.597   &  \textbf{24.779}       & 20.836            & 17.294          & 6.077           & 9                        & 11                  & 4                    \\ \hline
mimykgcp      & \textbf{25.082}       & \textbf{25.843}   & 8.536         & 18.691            & 23.929          & 7.999           & 11                       & \textbf{3}          & 4                    \\ \hline
DuyHUYNH      & 11.894     & 15.468     & 22.016       & 17.015            & 11.304          & \textbf{4.807}           & 13                       & 13                  & 4                    \\ \hline
 \hline
\textbf{baselines} & & & &  & & \\ \hline
 \hline
YOLOv3        & 13.249       & 17.176  & 20.617          & 17.374            & 15.456          & \textbf{4.397}           & 12                       & 12                  & \textbf{3}           \\ \hline
\begin{tabular}[c]{@{}l@{}}RetinaNet\\(ResNet101)\end{tabular}  & 8.754         & 8.079   &16.601          & 11.690            & 7.763           & \textbf{1.985}  & 14                       & 14                  & \textbf{3}           \\ \hline
\end{tabular}
}
}
\end{table}
%%%%%
%%%%%
%%%%%
\begin{table}[t!]
\centering
\small{
\caption{\textbf{Evaluation of the artefact segmentation task.} Top three best scores for each metric criteria are in bold.~\label{tab:ead2020_segmentation}}
\resizebox{\columnwidth}{!}{
\begin{tabular}{l|l|l|l|l|l|l|l|c} 
\hline
\centering
\begin{tabular}[c]{@{}l@{}}\textbf{Team }\\\textbf{Names}\end{tabular} & \textbf{JC} & \textbf{DSC} & \textbf{F2} & \textbf{PPV} & \textbf{Rec} & \textbf{Acc} & \textbf{Score$_s$} & \textbf{R$_{{score}_s}$}  \\ 
\hline \hline
qzheng5                                                                & {0.477}      & 0.532        & 0.561       & 0.556        & \textbf{0.835}        & 0.973        & 0.621             & 8                 \\ 
\hline
VinBDI                                                                 & \textbf{0.628}       & \textbf{0.673}        & \textbf{0.670}       & \textbf{0.837}        & 0.738        & \textbf{0.978}        & \textbf{0.730}             & \textbf{2}                 \\ 
\hline
higersky                                                               & 0.529       & 0.579        & 0.587       & 0.675        & 0.758        & 0.975        & 0.650             & 5                 \\ 
\hline
anand\_subu                                                              & 0.304       & 0.354        & 0.361       & 0.430        & 0.747        & 0.975        & 0.473             & 14                \\ 
\hline
arnavchavan04                                                          & \textbf{0.622}       & \textbf{0.673}        & \textbf{0.683}       & \textbf{0.800}        & 0.767        & \textbf{0.977}        & \textbf{0.731}             & \textbf{1}                 \\ 
\hline
DuyHUYNH                                                               & 0.502       & 0.557        & 0.583       & 0.593        & \textbf{0.829}        & 0.974        & 0.640             & 6                 \\ 
\hline
mimykgcp                                                               & 0.531       & 0.576        & 0.579       & \textbf{0.723}        & 0.726        & \textbf{0.977}        & 0.651             & 4                 \\ 
\hline
mouradai\_{ox}                                                             & \textbf{0.581}       & \textbf{0.632}        & \textbf{0.647}       & 0.711        & \textbf{0.800}        & 0.974        & \textbf{0.697}             & \textbf{3}                 \\ 
\hline
 \hline
\multicolumn{9}{l}{\textbf{baselines} }                                                                                                                                                                \\ 
\hline \hline
FCN8                                                                   & 0.500       & 0.548        & 0.550       & 0.670        & 0.708        & 0.976        & 0.619             & 9                 \\ 
\hline
UNet-ResNet34                                                           & 0.310       & 0.364        & 0.373       & 0.419        & 0.766        & 0.974        & 0.481             & 13                \\ 
\hline
PSPNet                                                                 & 0.497       & 0.541        & 0.534       & 0.698        & 0.680        & 0.975        & 0.613             & 10                \\ 
\hline
 \begin{tabular}[c]{@{}l@{}}DeepLabv3\\(ResNet50)                                                                                    \end{tabular}                                                  & 0.448       & 0.495        & 0.492       & 0.599        & 0.704        & 0.974        & 0.572             & 12                \\ 
\hline  
   \begin{tabular}[c]{@{}l@{}}DeepLabv3+\\(ResNet50)\end{tabular}                                                                                                 & 0.485       & 0.533        & 0.535       & 0.646        & 0.726        & 0.976        & 0.610             & 11                \\ 
\hline
\begin{tabular}[c]{@{}l@{}}DeepLabv3+\\(ResNet101)\end{tabular}                                                  & 0.501       & 0.547        & 0.546       & 0.683        & 0.718        & 0.973        & 0.624             & 7                 \\
\hline
\end{tabular}
}
}
\end{table}

\begin{figure}[t!]
    \centering
    \includegraphics[width= 1.0\textwidth]{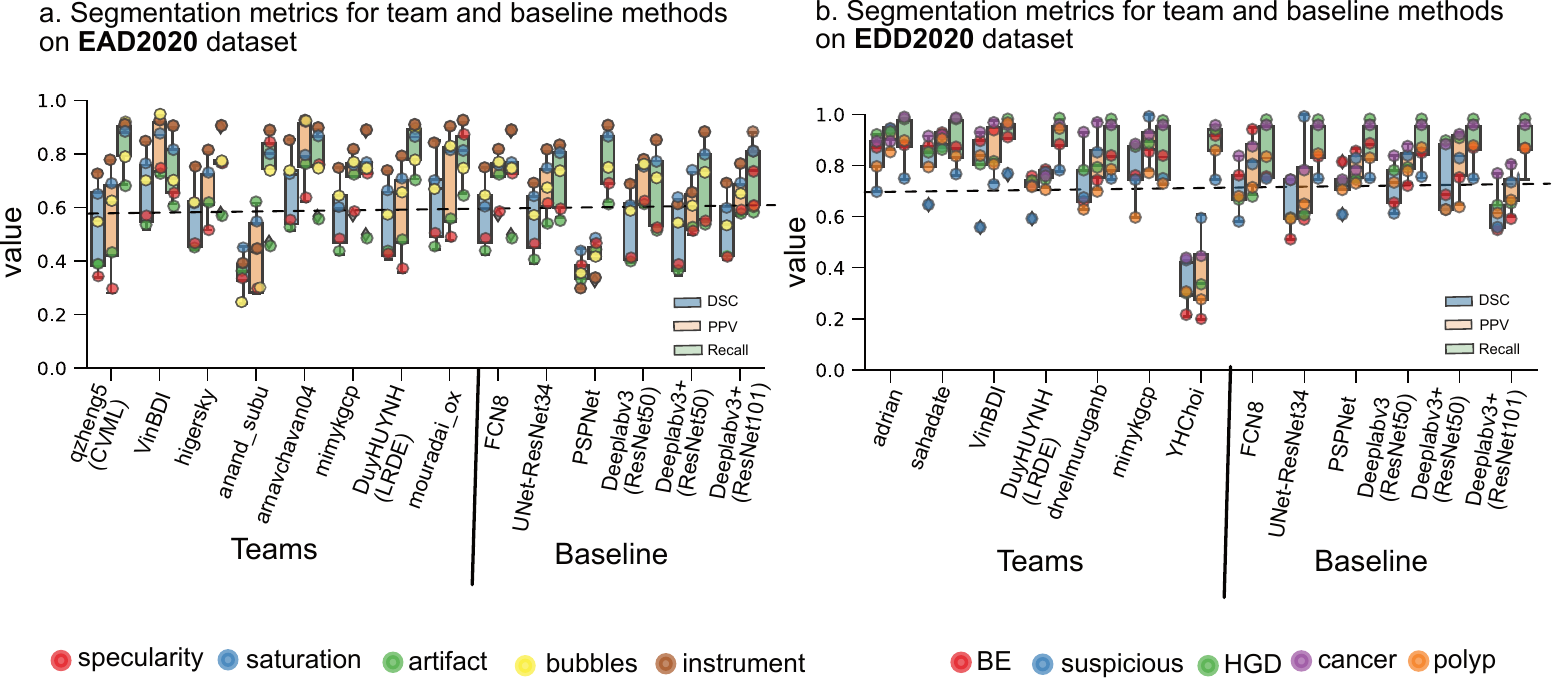}
    \caption{\textbf{Semantic segmentation for EAD and EDD sub-challenges}: Error bars with overlayed swarm plots for dice similarity coefficient (DSC), positive predictive value (PPV) or precision  and recall are presented for each team and baseline methods for the EAD2020 (a) and EDD2020 (b) challenges. 6 different baseline methods are also provided for comparison.}
    \label{fig:ead2020_segmentation_teams_comparison}
\end{figure}

\paragraph{Segmentation task for EAD2020}
%
%For the segmentation task, we compare the results of the participants w.r.t baseline segmentation methods: \textit{FCN8, UNet-ResNet34, PSPNet, DeepLabv3(ResNet50), DeepLabv3+(ResNet50)}, and \textit{DeepLabv3+(ResNet101)}. Table~\ref{tab:ead2020_segmentation} presents the JC, DSC, F2, PPV, recall, and accuracy obtained by each team and baseline methods when segmenting the different artefacts. As shown, the method proposed by team \textit{arnavchavan04} and team \textit{VinBDI} had the best performance in terms of JC, DSC, F2 and PPV proving the ability to segment less false positive regions. While the team \textit{arnavchavan04} ranked the first place with a score of 0.731 and accuracy of 0.977, the team \textit{VinBDI} was placed on 2nd place with an overall accuracy of 0.978 and a semantic score of 0.730. Both teams had an equal performance for the DSC with a value of 0.673. However, the method suggested by team \textit{qzheng5} and team \textit{DuyHUYNH} segmented more true positive regions compared to other teams obtaining top recall values of 0.8352 and 0.828. The baseline methods showed a low performance in terms of final score compared to the methods proposed by the participants coming to the end of the ranking list. In conclusion, the method proposed by team \textit{arnavchavan04} outperformed (i.e., R$_{{score}_s}$=1) in the overall performance for the different measures. 
%
{Table~\ref{tab:ead2020_segmentation} presents the JC, DSC, F2, PPV, recall, and accuracy obtained by each team and baseline methods. As shown, the method proposed by team \textit{arnavchavan04} and team \textit{VinBDI} had the best performance in terms of JC ($>$ 62\%), DSC ($>$ 67\%), F2 ($>$ 67\%) and PPV ($>$ 80\%) proving the ability to segment less false positive regions. However, the method suggested by team \textit{qzheng5} and team \textit{DuyHUYNH} segmented more true positive regions compared to other teams obtaining top recall values of 0.8352 and 0.828. The baseline methods showed a low performance in terms of final score compared to the methods proposed by the participants. Furthermore, Figure~\ref{fig:ead2020_segmentation_teams_comparison} (a) shows class-wise scores for DSC, PPV and Recall. Similar to detection, segmenting larger instances like the saturation and the instrument obtained the high scores.  Specularity, bubble and the artefact classes were among least performing classes for many teams and baseline methods.}

%While the team \textit{arnavchavan04} ranked the first place with a score of 0.731 and accuracy of 0.977, the team \textit{VinBDI} was placed on 2nd place with an overall accuracy of 0.978 and a semantic score of 0.730. Both teams had an equal performance for the DSC with a value of 0.673. However, the method suggested by team \textit{qzheng5} and team \textit{DuyHUYNH} segmented more true positive regions compared to other teams obtaining top recall values of 0.8352 and 0.828. The baseline methods showed a low performance in terms of final score compared to the methods proposed by the participants coming to the end of the ranking list. In conclusion, the method proposed by team \textit{arnavchavan04} outperformed (i.e., R$_{{score}_s}$=1) in the overall performance for the different measures. 
%
%
%Furthermore, Figure~\ref{fig:ead2020_segmentation_teams_comparison} (a) presents the scores for DSC, PPV and Recall for each class by the teams and baseline methods. Segmenting the saturation maintained the high score among the teams and the baseline methods for the three measures.  For the different teams, the highest recall was obtained by the specularity class while the lowest was shown when segmenting the artifact class. Moreover, similar to detection results, segmenting the bubble category showed a low performance for these three measures. 
%
\subsubsection{EDD2020 sub-challenge}
In this section, we report the performance of the participating teams in the EDD2020 challenge for the detection and segmentation.
\begin{table}[t!]
\centering
\small{
\caption{\textbf{EDD2020 results for the detection task on the single frame dataset.} mAP at IoU thresholds 25\%, 50\% and 75\% are provided along with overall mAP and overall IoU computations. Overall scores are computed at 11 IoU thresholds and averaged. Weighted detection score $score_d$ is computed between overall mAP and IoU scores only. Three best scores for each metric criteria are in bold.~\label{tab:edd2020_detection_singleframe}}
\resizebox{\columnwidth}{!}{
\begin{tabular}{l|c|c|c|c|c|c|c}
\hline
\multicolumn{1}{c|}{\textbf{\begin{tabular}[c]{@{}c@{}}Team \\ names\end{tabular}}} & \textbf{mAP$_{25}$} & \textbf{mAP$_{50}$} & \textbf{mAP$_{75}$} & \textbf{\begin{tabular}[c]{@{}c@{}}overall \\mAP$_{d}$\end{tabular}} & \textbf{\begin{tabular}[c]{@{}c@{}}overall\\ mIoU$_{d}$\end{tabular}} & \textbf{mAP$_\delta$} & \textbf{score$_d \pm \delta$ }  \\ \hline
\multicolumn{1}{l|}{adrian}                   & \multicolumn{1}{c|}{{  \textbf{48.402}}}     & \multicolumn{1}{c|}{{  \textbf{33.562}}}     & \multicolumn{1}{c|}{{ \textbf{27.098}}}     & \multicolumn{1}{c|}{{  \textbf{37.594}}}        & \multicolumn{1}{c|}{{ \textbf{ 27.614}}}         & \multicolumn{1}{c|}{8.523}            & \multicolumn{1}{c}{ \textbf{33.602 $\pm$ 8.523}}  \\ \hline
\multicolumn{1}{l|}{sahadate}                 & \multicolumn{1}{c|}{ \textbf{37.612}}   & \multicolumn{1}{c|}{ \textbf{23.284}}           & \multicolumn{1}{c|}{ \textbf{15.837}}           & \multicolumn{1}{c|}{ \textbf{26.834}}              & \multicolumn{1}{c|}{{  \textbf{32.420}}}          & \multicolumn{1}{c|}{8.325}            & \multicolumn{1}{c}{ \textbf{29.068 $\pm$ 8.325}}  \\ \hline
\multicolumn{1}{l|}{VinBDI}                   & \multicolumn{1}{c|}{{ \textbf{43.202}}}     & \multicolumn{1}{c|}{ \textbf{26.981}}           & \multicolumn{1}{c|}{{ \textbf{17.001}}}     & \multicolumn{1}{c|}{{ \textbf{30.219}}}        & \multicolumn{1}{c|}{17.773}               & \multicolumn{1}{c|}{ \textbf{9.478}}            & \multicolumn{1}{c}{25.241 $\pm$ 9.478}  \\ \hline
\multicolumn{1}{l|}{YHChoi}                   & \multicolumn{1}{c|}{23.183}           & \multicolumn{1}{c|}{11.082}           & \multicolumn{1}{c|}{8.800}              & \multicolumn{1}{c|}{15.783}              & \multicolumn{1}{c|}{24.623}               & \multicolumn{1}{c|}{6.216}            & \multicolumn{1}{c}{19.319 $\pm$ 6.216}  \\ \hline
\multicolumn{1}{l|}{DuyHUYNH}                 & \multicolumn{1}{c|}{23.959}           & \multicolumn{1}{c|}{9.587}            & \multicolumn{1}{c|}{5.659}            & \multicolumn{1}{c|}{12.479}              & \multicolumn{1}{c|}{13.829}               & \multicolumn{1}{c|}{6.284}            & \multicolumn{1}{c}{13.019 $\pm$ 6.284}  \\ \hline
\multicolumn{1}{l|}{mimykgcp}                & \multicolumn{1}{c|}{34.884}           & \multicolumn{1}{c|}{20.982}           & \multicolumn{1}{c|}{4.463}            & \multicolumn{1}{c|}{20.742}              & \multicolumn{1}{c|}{2.270}                 & \multicolumn{1}{c|}{ \textbf{9.359}}            & \multicolumn{1}{c}{13.353 $\pm$ 9.359}  \\ \hline
\multicolumn{1}{l|}{drvelmuruganb}            & \multicolumn{1}{c|}{31.018}           & \multicolumn{1}{c|}{18.421}           & \multicolumn{1}{c|}{11.768}           & \multicolumn{1}{c|}{21.790}               & \multicolumn{1}{c|}{7.322}                & \multicolumn{1}{c|}{7.424}            & \multicolumn{1}{c}{16.002 $\pm$ 7.424}  \\ \hline
 \hline
\textbf{baselines}                             & \multicolumn{1}{l}{}                  & \multicolumn{1}{l}{}                  & \multicolumn{1}{l}{}                  & \multicolumn{1}{l}{}                     & \multicolumn{1}{l}{}                      & \multicolumn{1}{l}{}                  & \multicolumn{1}{l}{}                     \\ \hline
 \hline
\multicolumn{1}{l|}{YOLOv3}                   & \multicolumn{1}{c|}{34.305}           & \multicolumn{1}{c|}{21.227}           & \multicolumn{1}{c|}{14.650}            & \multicolumn{1}{c|}{22.980}               & \multicolumn{1}{c|}{24.351}               & \multicolumn{1}{c|}{6.456}            & \multicolumn{1}{c}{23.528 $\pm$ 6.456}  \\ \hline
\multicolumn{1}{l|}{RetinaNet (ResNet50)}              & \multicolumn{1}{c|}{26.833}           & \multicolumn{1}{c|}{14.441}           & \multicolumn{1}{c|}{9.907}            & \multicolumn{1}{c|}{17.552}              & \multicolumn{1}{c|}{25.580}                & \multicolumn{1}{c|}{6.464}            & \multicolumn{1}{c}{20.763 $\pm$ 6.464}  \\ \hline
\multicolumn{1}{l|}{RetinaNet (ResNet101)}             

& \multicolumn{1}{c|}{42.579}           & \multicolumn{1}{c|}{{27.000}}       & \multicolumn{1}{c|}{11.194}           & \multicolumn{1}{c|}{27.974}              & \multicolumn{1}{c|}{ \textbf{26.434}}               & \multicolumn{1}{c|}{ \textbf{11.949}}           & \multicolumn{1}{c}{ \textbf{27.358 $\pm$ 11.949}} \\ \hline
\end{tabular}
}
}
\end{table}
\paragraph{Detection task for EDD2020}

{In Table~\ref{tab:edd2020_detection_singleframe}, the team \textit{adrian} achieved the highest score among other participants and the baseline methods with a final $score_d$ of {33.602 $\pm$ 8.523} with the highest overall mAP (37.594) and the second highest overall mIoU (27.614). The best localization score was obtained by the team \textit{sahadate} but with nearly 5\% lower mAP than the top scorer team. Furthermore, the baseline method~\textit{RetinaNet} with the ResNet101 backbone performed better than most of the participating teams. From Table~\ref{tab:edd2020_detection_perclass}, it is evident that most teams and baselines failed to detect suspicious class instance while most teams performed comparatively better on polyp and NDBE classes. Only the winning team \textit{adrian} and RetinaNet (ResNet101) provided a descent score for cancer class with most teams recording mAP below 10. For HGD class category, top performing teams were \textit{adrian} and\textit{VinBDI} with mAP over 25.}

\begin{table}[t!]
\centering
\small{
\caption{Per class evaluation results for the detection task of the EDD2020 sub-challenge.~\label{tab:edd2020_detection_perclass}}
\resizebox{\columnwidth}{!}{
\begin{tabular}{lcccccc}
\hline
\multicolumn{1}{c|}{\textbf{\begin{tabular}[c]{@{}c@{}}Teams \\ EDD2020\end{tabular}}} & \textbf{NDBE} & \textbf{suspicious} & \textbf{HGD} & \textbf{cancer} & \textbf{polyp} & \textbf{$\delta$}    \\ \hline
\multicolumn{1}{l|}{adrian}            & \multicolumn{1}{c|}{28.911} & \multicolumn{1}{c|}{ \textbf{1.776} }     & \multicolumn{1}{c|}{ \textbf{32.727}} & \multicolumn{1}{c|}{ \textbf{64.286}} & \multicolumn{1}{c|}{ \textbf{60.269}} & \multicolumn{1}{c}{ \textbf{22.841}} \\ \hline
\multicolumn{1}{l|}{sahadate}          & \multicolumn{1}{c|}{ \textbf{46.193}} & \multicolumn{1}{c|}{1.099}      & \multicolumn{1}{c|}{22.727} & \multicolumn{1}{c|}{10.000}   & \multicolumn{1}{c|}{54.152} & \multicolumn{1}{c}{ \textbf{20.414}} \\ \hline
\multicolumn{1}{l|}{VinBDI} & \multicolumn{1}{c|}{ \textbf{48.489}} & \multicolumn{1}{c|}{ \textbf{3.497}}      & \multicolumn{1}{c|}{ \textbf{25.852}} & \multicolumn{1}{c|}{10.000}   & \multicolumn{1}{c|}{ \textbf{63.260}}  & \multicolumn{1}{c}{ \textbf{22.660}}  \\ \hline
\multicolumn{1}{l|}{YHChoi}            & \multicolumn{1}{c|}{26.900}   & \multicolumn{1}{c|}{0.000}        & \multicolumn{1}{c|}{22.727} & \multicolumn{1}{c|}{0.000}    & \multicolumn{1}{c|}{29.289} & \multicolumn{1}{c}{13.057} \\ \hline
\multicolumn{1}{l|}{DuyHUYNH}          & \multicolumn{1}{c|}{20.281} & \multicolumn{1}{c|}{1.499}      & \multicolumn{1}{c|}{11.364} & \multicolumn{1}{c|}{0.000}    & \multicolumn{1}{c|}{29.254} & \multicolumn{1}{c}{11.134} \\ \hline
\multicolumn{1}{l|}{mimykgcp}         & \multicolumn{1}{c|}{ \textbf{50.089}} & \multicolumn{1}{c|}{ \textbf{4.592}}      & \multicolumn{1}{c|}{ \textbf{23.064}} & \multicolumn{1}{c|}{5.852}  & \multicolumn{1}{c|}{20.112} & \multicolumn{1}{c}{16.429} \\ \hline
\multicolumn{1}{l|}{drvelmuruganb}     & \multicolumn{1}{c|}{34.775} & \multicolumn{1}{c|}{0.000}        & \multicolumn{1}{c|}{22.727} & \multicolumn{1}{c|}{0.000}    & \multicolumn{1}{c|}{51.446} & \multicolumn{1}{c}{19.993} \\ \hline \hline
\textbf{baselines}                      & \multicolumn{1}{l}{}        & \multicolumn{1}{l}{}            & \multicolumn{1}{l}{}        & \multicolumn{1}{l}{}        & \multicolumn{1}{l}{}        & \multicolumn{1}{l}{}        \\ \hline
 \hline
\multicolumn{1}{l|}{YOLOv3 (darknet53)}            & \multicolumn{1}{c|}{38.839} & \multicolumn{1}{c|}{0.000}        & \multicolumn{1}{c|}{6.970}   & \multicolumn{1}{c|}{ \textbf{16.667}} & \multicolumn{1}{c|}{52.426} & \multicolumn{1}{c}{19.712} \\ \hline
\multicolumn{1}{l|}{RetinaNet (ResNet50) }       & \multicolumn{1}{c|}{23.636} & \multicolumn{1}{c|}{0.000}        & \multicolumn{1}{c|}{18.182} & \multicolumn{1}{c|}{0.000}    & \multicolumn{1}{c|}{45.943} & \multicolumn{1}{c}{17.086} \\ \hline
\multicolumn{1}{l|}{RetinaNet (ResNet101 )}      & \multicolumn{1}{c|}{29.483} & \multicolumn{1}{c|}{0.000}        & \multicolumn{1}{c|}{22.727} & \multicolumn{1}{c|}{ \textbf{31.818}} & \multicolumn{1}{c|}{ \textbf{55.840}}  & \multicolumn{1}{c}{17.909} \\ \hline
\end{tabular}
}
}
\end{table}
\paragraph{Segmentation task for EDD2020}
\begin{table}[b!]
\centering
\small{
\caption{\textbf{Evaluation of the disease segmentation methods proposed by the participating teams and the baseline methods.} Top three evaluation criteria are highlighted in bold.~\label{tab:edd2020_segmentation_singleframe}}
\resizebox{\columnwidth}{!}{
\begin{tabular}{l|l|l|l|l|l|l|l|c} 
\hline
\begin{tabular}[c]{@{}l@{}}\textbf{Team}\\\textbf{Names}\end{tabular} & \textbf{JC} & \textbf{DSC} & \textbf{F2} & \textbf{PPV} & \textbf{Rec} & \textbf{Acc} & \textbf{Score$_s$} & \textbf{R$_{{score}_s}$}  \\ 
\hline
adrian                                                                        &  \textbf{0.820}      &  \textbf{0.836 }       &  \textbf{0.842 }     &  \textbf{0.921 } & 0.894       & 0.955       & \textbf{ 0.873}                &  \textbf{1  }                   \\ 
\hline
sahadate                                                                      & \textbf{ 0.797 }     &  \textbf{0.816 }      &  \textbf{0.819}     &  \textbf{0.906 }      & 0.883       & 0.955       &  \textbf{0.856 }            &  \textbf{2}                     \\ 
\hline
VinBDI   & \textbf{ 0.788 }     &  \textbf{0.805}       &  \textbf{0.812 }     &  \textbf{0.859}       &  \textbf{0.912}       & 0.952       &  \textbf{0.847   }       &  \textbf{3}                     \\ 
\hline
DuyHUYNH                                                                      & 0.6843      & 0.7058       & 0.718     & 0.762        &  \textbf{0.905}      & 0.931       & 0.773                 & 9                     \\ 
\hline
drvelmuruganb                                                                 & 0.7166      & 0.7349       & 0.734      & 0.819       & 0.857      & 0.959      & 0.786               & 6                     \\ 
\hline
mimykgcp                                                                     & 0.7561      & 0.7721       & 0.770      & 0.893       & 0.845       & 0.957       & 0.820               & 4                     \\ 
\hline
YHChoi                                                                        & 0.314      & 0.340       & 0.356       & 0.385       & \textbf{ 0.896}      & 0.892       & 0.494                & 13                    \\ 
\hline
 \hline
\multicolumn{9}{l}{\textbf{baselines}}                                                                                                                                                                                \\ 
\hline
\hline
FCN8                                                                          & 0.687      & 0.705       & 0.709      & 0.811       & 0.850       & 0.953       & 0.769                & 10                    \\ 
\hline
UNet-ResNet34                                                                 & 0.617      & 0.637       & 0.638      & 0.732      & 0.868       & 0.958       & 0.719                & 11                    \\ 
\hline
pspnet                                                                        & 0.698      & 0.721       & 0.723      & 0.797       & 0.876       & 0.959       & 0.779              & 8                     \\ 
\hline
\begin{tabular}[c]{@{}l@{}}DeepLabv3\\(RetinaNet50)\end{tabular}              & 0.704      & 0.724       & 0.724      & 0.810       & 0.878       &  \textbf{0.962 }      & 0.784              & 7                     \\ 
\hline
\begin{tabular}[c]{@{}l@{}}DeepLabv3+\\(RetinaNet50)\end{tabular}             & 0.725     & 0.744       & 0.749      & 0.818       & 0.882       &  \textbf{0.960 }      & 0.798                & 5                     \\ 
\hline
\begin{tabular}[c]{@{}l@{}}DeepLabv3+\\(RetinaNet1010\end{tabular}            & 0.608       & 0.627        & 0.629     & 0.698       & 0.880       &  \textbf{0.962 }      & 0.709               & 12                   \\ \hline
\end{tabular}
}
}
\end{table}
%
%The results of the participant teams for disease segmentation from endoscopic frames are reported in this section. In table~\ref{tab:edd2020_segmentation_singleframe}, we represent the JC, DSC, F2, PPV, recall and accuracy obtained by each team for the segmentation task. As illustrated, three teams \textit{(Adrian, sahadate} and \textit{nhanthanhnguyen94}) achieved a high overlapping ratio between the generated masks and the ground-truth annotations with a JC value of 0.8203, 0.798, and 0.7882, respectively. Moreover, they maintained the high performance for the DSC, F2, and PPV measures with comparable results as well.
%
%Also, the method presented by the teams \textit{VinBDI, DuyHUYNH} and \textit{YHChoi} were able to segment more true positive regions reaching the top recall values of 0.912, 0.905 and 0.896, respectively. The overall accuracy result for all the participating teams and the baseline methods were very competitive.
%  
%Additionally, the scores for DSC, PPV, and recall for each class by the teams and baseline methods are shown in Fig.~\ref{fig:ead2020_segmentation_teams_comparison} b. Most participating teams and the baseline methods showed highest values for cancerous regions. Also, most teams showed higher DSC, PPV and recall for BE class instance as well ($> 0.8$ for top three teams). As presented, the majority of metrics had the least values for the suspicious class. Also, the segmentation of polyps did not show good performance in comparison to the detection results.

{From Table~\ref{tab:edd2020_segmentation_singleframe}, it can be observed that the three teams \textit{(Adrian, sahadate} and \textit{nhanthanhnguyen94}) achieved a DSC over 0.80. Moreover, they maintained the high performance for other metrics as well that include JC ($>$0.78), F2 ($>$0.81), and PPV ($>$0.85) securing first, second and third ranks, respectively. Teams \textit{VinBDI} and  \textit{DuyHUYNH} were able to segment more true positive regions reaching the top recall values. Fig.~\ref{fig:ead2020_segmentation_teams_comparison} (b) represents per-class metric values. It can be observed that unlike detection task, most teams reported high performance for cancer class. Also, most teams showed higher DSC, PPV and recall for BE class instance as well ($> 0.8$ for top three teams). However, similar to the detection task, most team and baseline methods reported least values for the suspicious class.}
  
%Additionally, the scores for DSC, PPV, and recall for each class by the teams and baseline methods are shown in Fig.~\ref{fig:ead2020_segmentation_teams_comparison} b. Most participating teams and the baseline methods showed highest values for cancerous regions. Also, most teams showed higher DSC, PPV and recall for BE class instance as well ($> 0.8$ for top three teams). As presented, the majority of metrics had the least values for the suspicious class. Also, the segmentation of polyps did not show good performance in comparison to the detection results.
%%%%%%%%%%%%%%%%%%%%%%%%%%%
\subsection{Qualitative results}
%%%%%%%%%%%%%%%%%%%%%%%%%%%
% Det
\begin{figure}[t!]
    \centering
    \includegraphics[width= 1.0\textwidth]{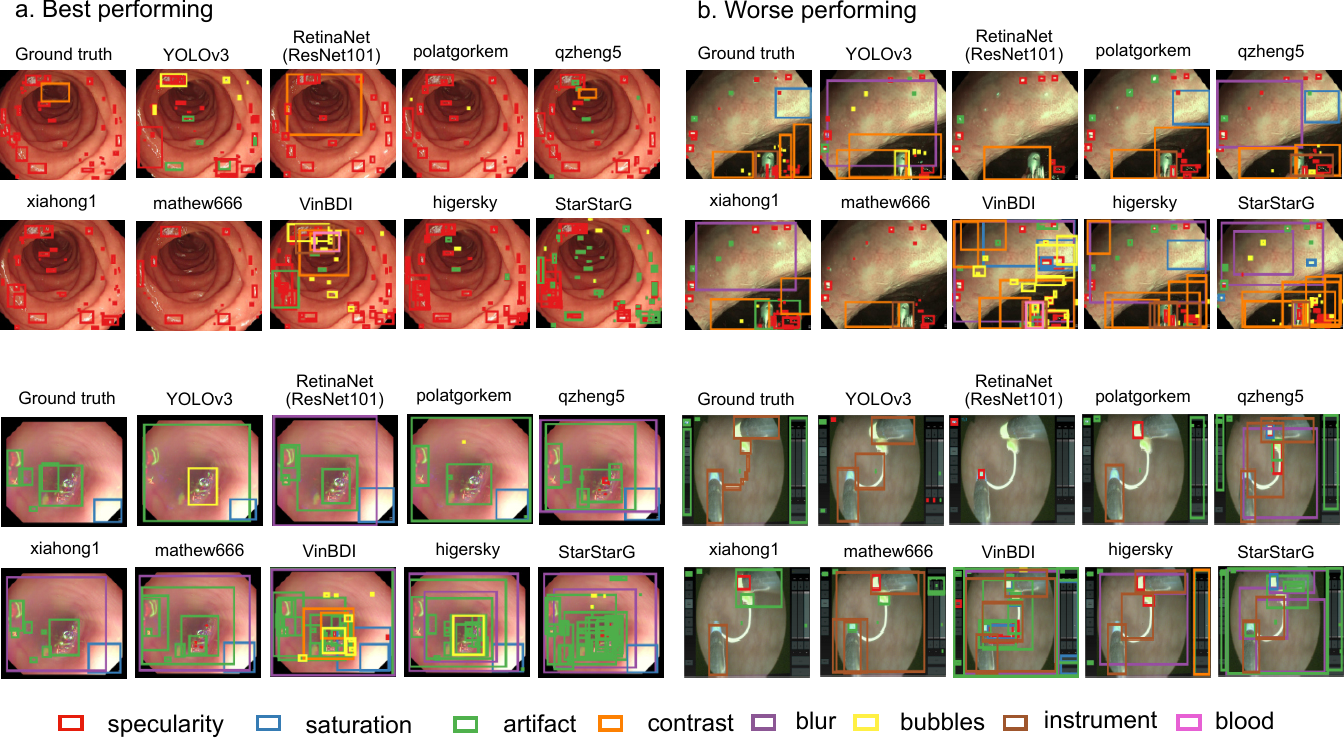}
    \caption{\textbf{EAD2020 best and worse performing samples for the detection task}. a) Best performing samples for 6 top ranked team results. b) Worse performing samples for the same teams in (a). Results with baseline methods are also included together with ground truth sample.}
    \label{fig:EAD_detection_results}
\end{figure}
% Det description
\subsubsection*{Detection task}
Figure~\ref{fig:EAD_detection_results} shows the best (panel a) and the worse (panel b) performing frames from single frame dataset for EAD2020. It can be observed that specularity and artefacts are detected and well localized by top teams (see Figure~\ref{fig:EAD_detection_results} a). Similarly, in the bottom example, saturation is also detected by all the participants. Even though, blur is not present for this sample, most methods also detected it. % explain in discussion
While for the worse performing frame (see Figure~\ref{fig:EAD_detection_results} b), instrument class is confused with contrast or artefact on the top sample, while in the bottom sample instrument is detected by some teams but often either detected only partially or overlapped by different classes such as saturation or artefact. 

%is either not detected or confused with artefact or saturation.
% Gen
% Gen-figure
\begin{figure}[t!]
    \centering
    \includegraphics[width=1.0\textwidth]{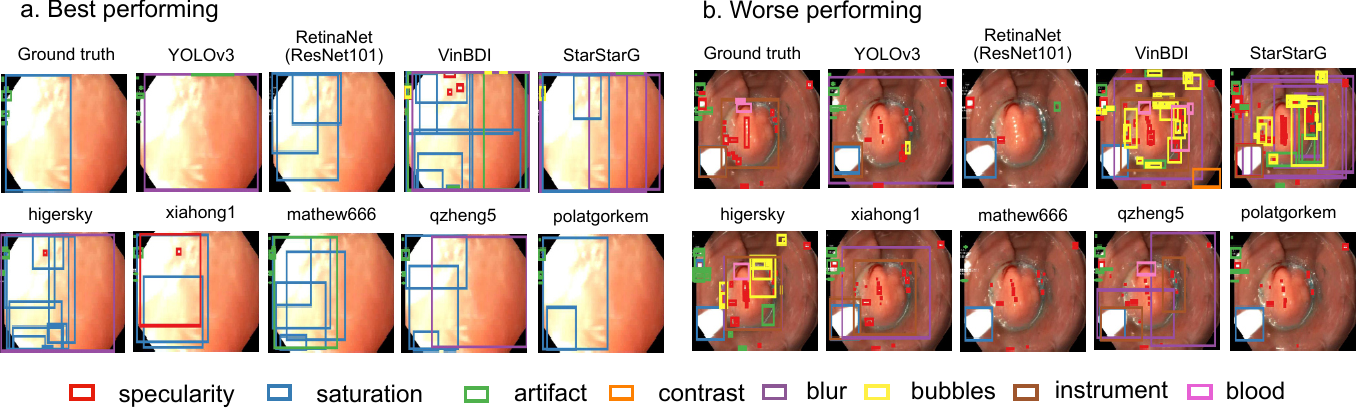}
    \caption{\textbf{EAD2020 best and worse performing samples for the generalization task}. a) Best performing samples for 7 top ranked team results. b) Worse performing samples for the same teams in (a). Results with baseline methods are also included together with ground truth sample.}
    \label{fig:EAD_genera_results}
\end{figure}
%Gen
For out-of-sample generalization task, it can be seen in Figure~\ref{fig:EAD_genera_results} (a) that besides YOLOv3 baseline method, all the baselines and teams detected saturation class. While some teams (\textit{mathew666}, \textit{VinBDI}, \textit{higersky}) detected multiple bounding boxes for the same class, they also detected blur class for this frame. While for worse performing frame (see Figure~\ref{fig:EAD_genera_results} (b)), instrument class (at the center of the image) is well localized only by the team~\textit{xiahong1} while most teams either partially detected the instrument (e.g., team \textit{qzheng5}) or could not detect the instrument class at all (e.g., team \textit{polatgorkem}). In both cases, the three teams \textit{VinBDI, higersky} and \textit{StarStarG} produced multiple overlapping and different size bounding boxes. 
% Det-figure
\begin{figure}[t!]
    \centering
    \includegraphics[width= 1.0\textwidth]{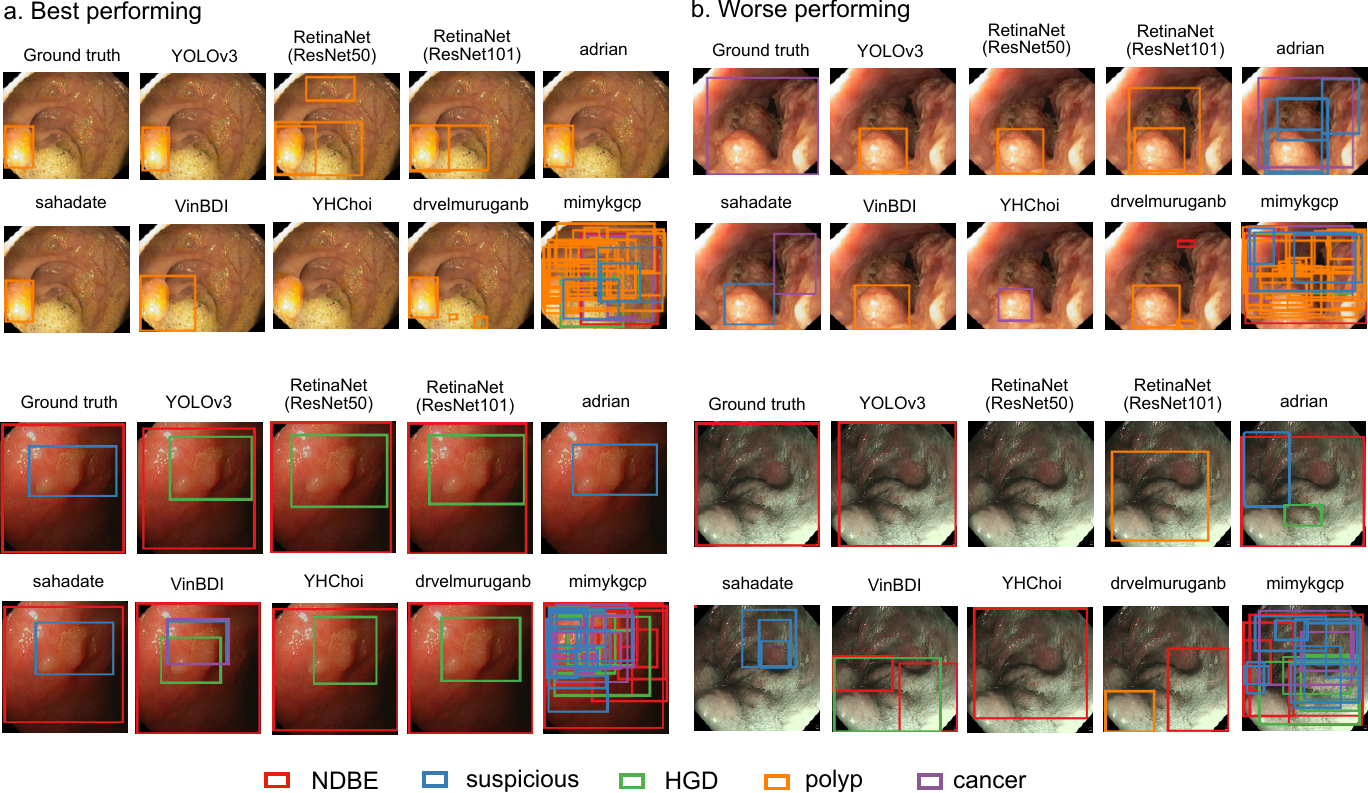}
    \caption{\textbf{EDD2020 best and worse performing samples for the detection task}. a) Best performing samples for 6 top ranked team results. b) Worse performing samples for the same teams in (a). Results with baseline methods are also included together with ground truth sample.}
    \label{fig:EDD_detection_results}
\end{figure}
% Segmentation
% Seg-EAD2020-Figure
\begin{figure}[t!]
    \centering
    \includegraphics[width= 1\textwidth]{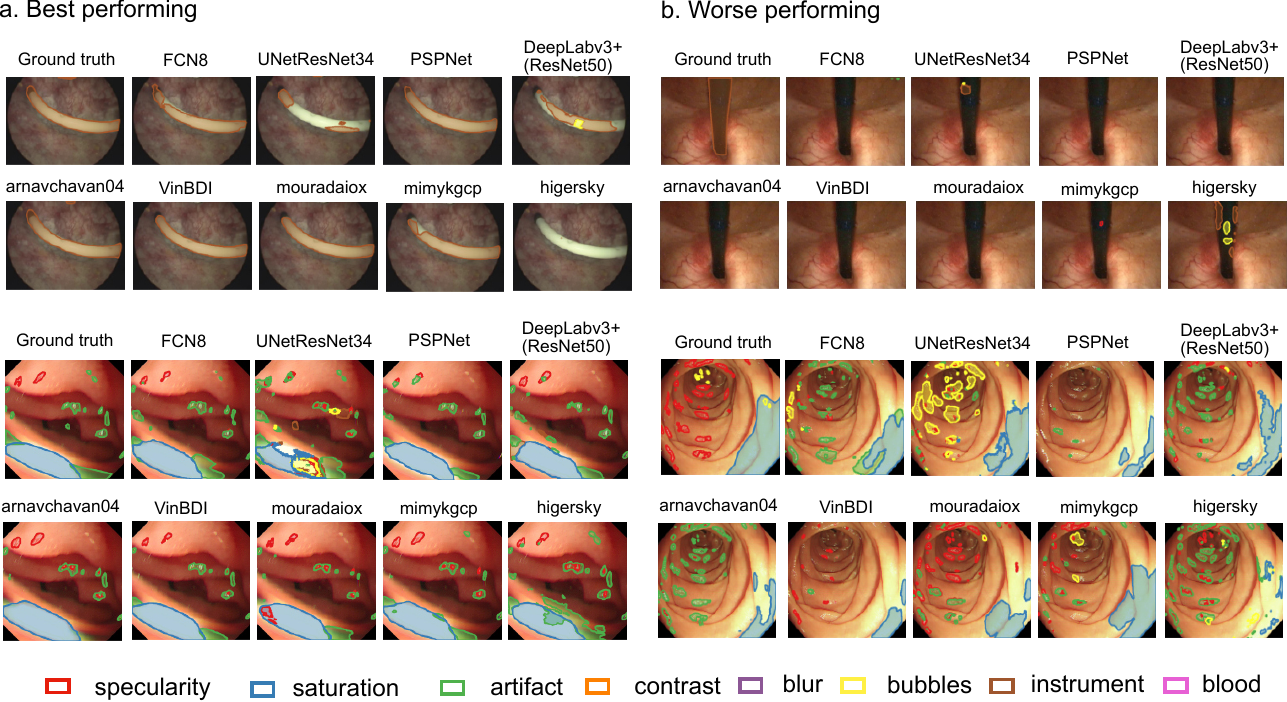}
    \caption{\textbf{EAD2020 best and worse performing samples}. a) Best performing samples for 5 top ranked team results. b) Worse performing samples for the same teams in (a). Results with baseline methods are also included together with ground truth sample (top). Single class samples are chosen at the top and multi-class samples are at the bottom in each category.}
    \label{fig:EAD_segmentation_results}
\end{figure}
\begin{figure}[t!]
    \centering
    \includegraphics[width= 0.98\textwidth]{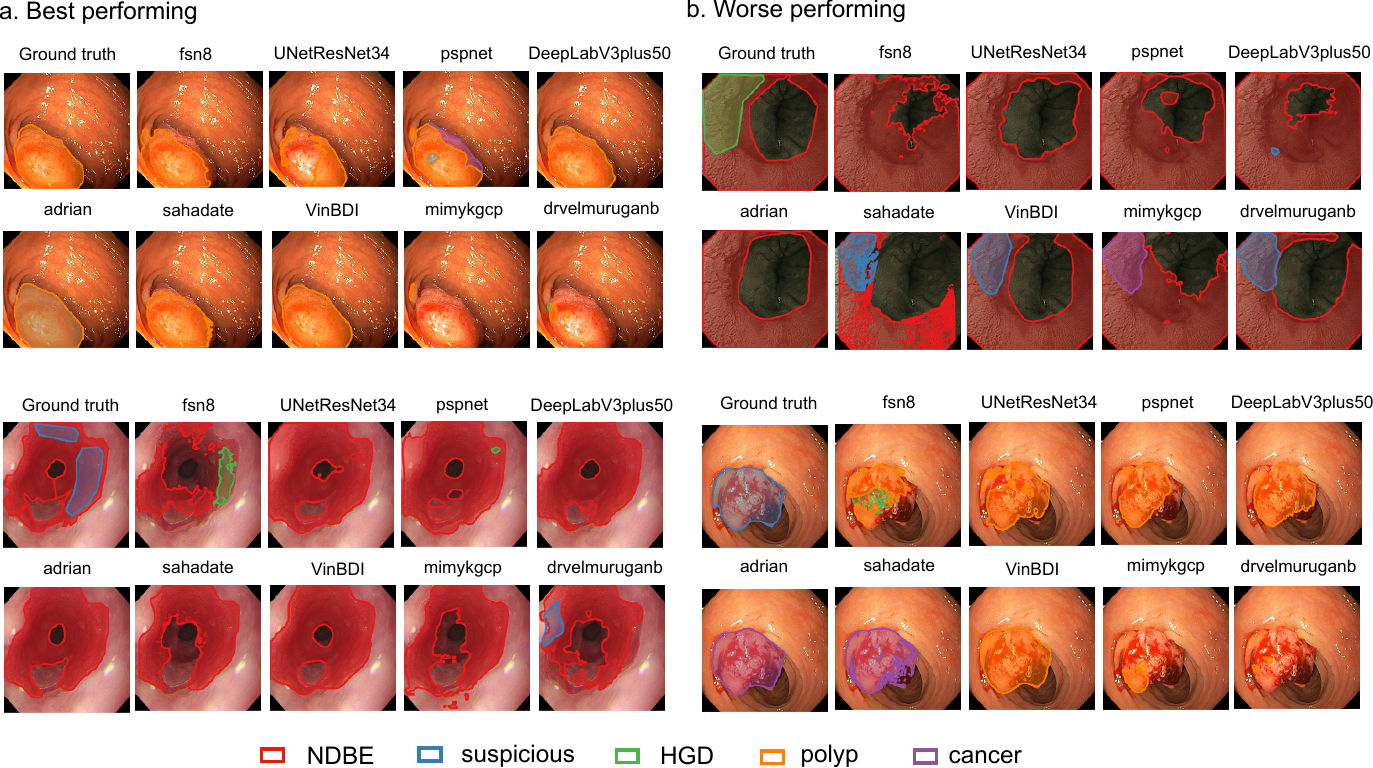}
    \caption{\textbf{EDD2020 best and worse performing samples}. a) Best performing samples for 5 top team results. b) Worse performing samples for the same teams in (a). Results with baseline methods are also included together with ground truth sample (top).}
    \label{fig:EDD_segmentation_results}
\end{figure}
Qualitative results for the EDD2020 challenge is shown in Figure~\ref{fig:EDD_detection_results}. The best performing samples in Figure~\ref{fig:EDD_detection_results} (a) shows polyp class (at the top); non-dysplastic Barrett's esophagus (NDBE) and suspicious classes on the bottom. It can be observed that polyp class is detected and well localized by all the teams and baseline methods. However, for bottom row NDBE is detected by most of the methods while confusion is observed across the suspicious class with high-grade dysplasia (HGD) class. Team \textit{mimykgcp} produced numerous bounding boxes failing to optimally localize adherent disease classes.
For the worse performing frames (Figure~\ref{fig:EDD_detection_results} (b)), cancer class (top) in the ground truth is confused with the polyp class instance for most of the teams and the baseline methods. While, for the NDBE class in the bottom of Figure~\ref{fig:EDD_detection_results} (b), teams were either not able to detect the NDBE class (except team \textit{adrian}, team \textit{YHChoi} and YOLOv3) at all or partially detected the NDBE areas  (e.g., teams \textit{VinBDI} and \textit{drvvelmuruganb}). Again, for the presented case, team \textit{mimykgep} detected numerous bounding boxes.  
%%%%%%%%%%%%%%%%%%%%%%%%%%%%
% Seg-EAD2020-description
%
\subsubsection*{Segmentation task}
Endoscopic artefact segmentation samples representing best and worse performing teams is provided in Figure~\ref{fig:EAD_segmentation_results}. For the sample with only the instrument class (see Figure~\ref{fig:EAD_segmentation_results} a, top panel) it can be observed that almost all the baseline and teams were able to predict precise delineation of the instrument class. Similarly, in the bottom panel of Figure~\ref{fig:EAD_segmentation_results} (a), specularity, saturation and artefact classes were segmented well by most of the teams and baseline methods. 
Even though, a single instrument class is present in the sample image in Figure~\ref{fig:EAD_segmentation_results} (b), none of the methods were able to segment the instrument. Also, for the bottom panel in the Figure~\ref{fig:EAD_segmentation_results} (b), specularity areas were segmented well by the teams \textit{mouradaiox} and \textit{ mimykgcp}. However, saturation area was under segmented by most of the teams and baseline methods. 
%
% Seg-EDD2020 figure
%
%% Seg-EDD2020 description
%
Figure~\ref{fig:EDD_segmentation_results} (a) represents the polyp class (at the top); NDBE and suspicious classes (at the bottom). It can be observed that polyp is segmented well by all the baselines and most teams (except team \textit{drvelmuruganb} who misclassified the pixels to suspicious class). While, most teams and baselines were able to precisely delineate NDBE class for the frame in the bottom panel but missed suspicious area.
In the worse performing sample (see Figure~\ref{fig:EDD_segmentation_results} (b)), most teams were able to segment NDBE area but large HGD area was missed by all the teams. Also, some teams confused HGD area with suspicious class.
For the bottom panel in Figure~\ref{fig:EDD_segmentation_results} (b), instead of suspicious class present in the ground truth, almost all the teams detected this as polyp or cancer. However, the region delineation was close to the ground truth for most teams. 

\section{Discussion}
\begin{figure}
    \centering
    \includegraphics[width= 1.0\textwidth]{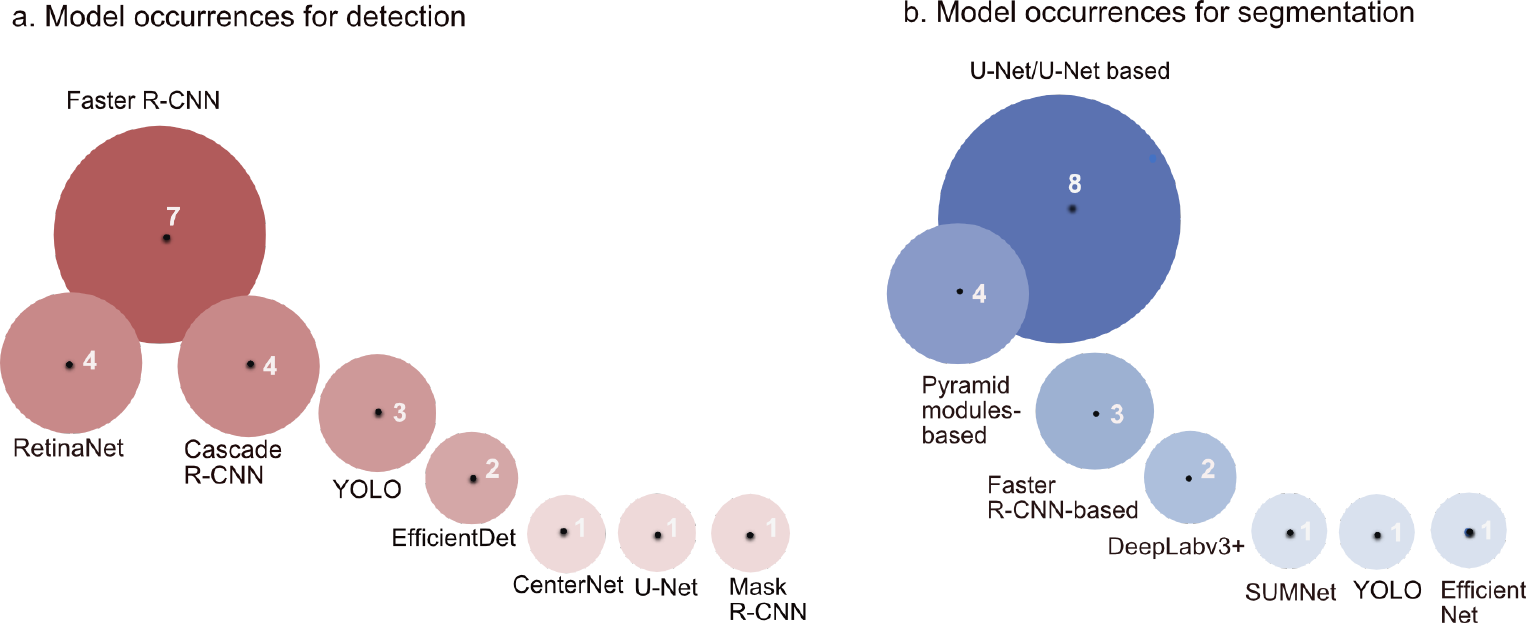}
    \caption{EndoCV2020 method categories in blob-representation. Model occurrences are presented for detection (a) and segmentation (b) tasks for both EAD2020 and EDD2020 sub-challenges. The number of occurrences is provided inside each blob.}
    \label{fig:method_blobs}
\end{figure}
% These will include discussions on what we participants have used, what are the best practices based on results and future direction.
% \begin{itemize}
%     \item Preprocessing effect
%     \item Loss function, model (highlight single stage or 2 stage or combined models) 
%     \item Generalization 
%     \item Performance (runtime) on GPU and CPU (if available)
%     \item Future directions
% \end{itemize}

Deep learning methods are rapidly being translated for the use of computer aided detection (CADe) and diagnosis (CADx) of diseases in complex clinical settings including endoscopy. However, the amount of data variability particularly in endoscopy is significantly higher than in natural scenes which possess a significant challenge in the process. It is therefore vital to determine an effective translational pathway in endoscopy. Majority of challenges in endoscopy are due to its complex surveillance that lead to severe artefacts that may confuse with disease. Similarly, a system designed for a particular organ may not generalize to be used in the other. 
%The challenge tackles well with 

Most deep learning methods that were used in the EndoCV2020 challenge can be categorised into multiscale, symbiotic, ensemble, encoder-decoder and cascading nature, or a combination of these (see Table~\ref{tab:EAD2020_method_summary} and Table~\ref{tab:EDD2020_method_summary}). Figure~\ref{fig:method_blobs} presents the overview of the used methods for the detection (a) and segmentation (b) challenge tasks based on the architecture usage. It can be observed that the majority of detection methods used two-stage Faster-RCNN with 4/7 teams combining it with one-stage RetinaNet or YOLOv3 or a combination of all. Cascade R-CNN which is built upon Faster R-CNN cascaded architecture was exploited by 4 teams. Similarly, U-Net-based architectures were utilised by most teams for semantic segmentation task with 4 teams exploring pyramid module-based architectures and 2 teams used Deeplabv3+ architecture. Faster RCNN-based model was also explored with additional thresholding (e.g., team \textit{adrian}) or per pixel prediction heads (e.g., team \textit{sahadate}). {Even though similar techniques were used in EAD2019 challenge~\cite{ali2020objective},  a direct comparison is not possible. This is due to the inclusion of more data for EAD2020 in both train and test sets. Also, EAD2020 includes sequence data which was not provided in EAD2019 challenge.}

For the detection task, the top performing teams on the challenge metric in both EAD (team \textit{polatgorkem}) and EDD (team \textit{adrian}) were those using ensemble networks, i.e., maneuvering outputs from multiple architectures. However, these networks sacrifice the speed of detection which can be observed from the computational time which were significantly higher than teams that used a single architecture (see Table~\ref{tab:ead2020_detection_ranking} and Table~\ref{tab:edd2020_detection_singleframe}). Other teams that used such an approach included team \textit{arnavchavan04} and \textit{mimykgcp} who combined Faster R-CNN with RetinaNet but both teams were respectively on 10th and 11th ranking. Just using Faster R-CNN alone with ResNet101 backbone, teams \textit{qzhang5} and \textit{mathew666} were able to detect both small and large size bounding boxes with sub-optimal accuracy that put them at 2nd and 4th positions, respectively. Similarly, team \textit{sahadate} claimed 2nd position on EDD detection task using Mask R-CNN which is based on the Faster R-CNN architecture. {For EAD2019 challenge~\cite{ali2020objective}, team yangsuhui also used an ensemble network with Cascade RCNN and FPN approach for the detection task similar to the EAD2020 top scorer team \textit{polatgorkem}.}

An intelligent choice for improved speed and accuracy using a scalable network was presented by the teams \textit{xiahong1} (used YOLACT) and \textit{VinBDI} (used EfficientDet D0) which were placed 3rd and 5th, respectively, on the final detection score of the EAD2020. {On the sequence data, team \textit{VinBDI} was the 2nd best method demonstrating the reliability of the used EfficientNet and FPN architectures. However, for almost all team methods the standard deviation was higher than for single frame data. No team exploited the sequence data provided for training.} Team \textit{VinBDI} was also ranked 3rd on the EDD detection task. Teams \textit{higerssky}, \textit{StarStarG} and \textit{MXY} that used cascaded R-CNN were ranked respectively on 6th, 7th and 9th positions. Additionally, the team StarStarG was ranked 1st and team higersky was ranked 2nd on the overall mAP.
However, it is to be noted that taking only mAP scores into account for detection could lead to over detection of the bounding boxes that increases the chance of finding a particular class but at the same time weakens the localization capability of the algorithm (see Figure~\ref{fig:EAD_detection_results}). Similar observations were found for the EDD dataset where the team \textit{mimykgcp} obtained an overall mAP of 20.742 but only 2.270 for the overall IoU (see Table~\ref{tab:edd2020_detection_singleframe}). As a result, over detection of the bounding boxes can be seen in Figure~\ref{fig:EDD_detection_results}.
In order to deal with the over detection of the bounding boxes, YOLACT architecture used by \textit{xiahong1} suppressed the duplicate detections using already-removed detections in parallel (\textit{fast NMS}). Similarly, teams such as \textit{polatgorkem} from the EAD and \textit{adrian} from the EDD were able to eliminate the duplicate detections using ensemble network and a class agnostic NMS.

\textit{Hypothesis I: In the presence of multiple class objects, object detection methods may fail to precisely regress the bounding boxes. Methods need better penalisation on the bounding box regression or a technique to perform effective non-maximal suppression.}

The choice of networks from each team depended on their ambition of either obtaining very high accuracy without focusing on speed or a trade-off between the speed and the accuracy or focusing on both and thinking out-of the box to use more recent developed methods which beats faster networks (such as YOLOv3) that included EfficientDet D0 architecture used by the team VinBDI (see Table \ref{tab:EAD2020_method_summary}).
% For detection generalisation who was good? Why?
Due to the efficiency of the EfficientDet D0 network that used biFPN and efficientNet backbone, team \textit{VinBDI} achieved second least deviation in mAP (i.e., dev$_g$ $=5.607$) with competitive mAP$_g$ ($=24.140$) and won the generalization task together with the team \textit{StarStarG} who had slightly higher mAP$_g$ ($=25.340$) but larger mAP deviation between detection and generalization datasets.
Most methods for the detection task on both the EAD and EDD dataset performed better than the baseline one-stage methods (YOLOv3 and RetinaNet). However, it was found that even though team \textit{polatgorkem} won the detection task, the method failed on generalization data where the team was ranked only last. The main reason behind this could be because the generalization gap mAP$_g$ was estimated between two mAP's (mAP$_d$ and mAP$_g$) and not IoU. Also, the final ranking was done taking into account the rank of dev$_g$ and mAP$_g$ only. It can be observed in Figure~\ref{fig:EAD_genera_results} that the bounding box localization of team \textit{polatgorkem} is precise in (a) while it misses instrument area at the center in (b). However, the winning teams \textit{VinBDI} and \textit{StarStarG} both over detect the boxes. The generalization ability of the methods were not explored for EDD dataset.

\textit{Hypothesis II: Metrics are critical but using a single metric does not always gives the right answer. Weighted metrics are desired in object detection task to establish a good trade-off between detection and precise localization.}

A major problem in the detection of EDD dataset was class confusion mostly for suspicious, HGD and cancer classes. This could be because of smaller number of samples for each of these classes compared to NDBE and polyp (see Figure~\ref{fig:dataset_samples}). While most methods were able to detect and localize NDBE and polyp class in general (3/7 teams with an overall mAP $>45$ and 4/7 teams with $>50$), all teams failed in suspicious class (overall mAP $<5.0$) and most teams for cancer class (overall mAP $<15.0$) (see Table~\ref{tab:edd2020_detection_perclass}). Figure~\ref{fig:EDD_detection_results} shows that polyp is detected and localized very well by most teams (a, top). Similarly, NDBE is localized by most methods, however, in this case suspicious class is confused mostly with the HGD. Also, in Figure~\ref{fig:EDD_detection_results} (b, top), it can be observed that the cancer class instance is confused with mostly polyp class.

\textit{Hypothesis III: Detection bounding boxes confuse with classes that have similar morphology and smaller number of samples failing to learn the contextual features. To improve detection, such samples need to be identified and more data demonstrating such attributes need to be injected (both positive and negative samples).}
%

% Segmentation
%
Similar to the detection task, teams that used ensemble techniques were among the best performing teams for the segmentation task. Teams \textit{arnavchavan04} and \textit{VinBDI} secured first (\textit{score}$_s=0.731$) and second (\textit{score}$_s=0.730$) positions, respectively, on the EAD2020 segmentation task (see Table~\ref{tab:ead2020_segmentation}) and the team \textit{adrian} won the EDD2020 segmentation task challenge with \textit{score}$_s$ of 0.873 (see Table~\ref{tab:edd2020_segmentation_singleframe}). The team \textit{arnavchavan04} used multiple augmentation techniques including cutmix and a feature pyramid network with a combination of EfficientNet backbones from B3 to B5. Similarly, team \textit{VinBDI} ensembled a U-Net architecture with EfficientNet B4 and BiFPN network with ResNet50 backbone. 
{Compared to EAD2019 where the winning team yangsuhui used DeepLabV3+ model with two different backbones, both of the top scorer teams of 2020 revealed the strength of recent EfficientNet and FPN-based segmentation approaches.}

In the EDD2020 segmentation task, the team \textit{adrian} combined predictions from three object detection architectures where the YOLOv3 and Faster R-CNN class predictions were used to correct the instance segmentation masks from Cascade R-CNN. A direct instance segmentation approach used by the team \textit{sahadate} secured second position (\textit{score}$_s$ = 0.856) on the same while ensemble network of the team \textit{VinBDI} secured the third position (\textit{score}$_s$ = 0.847). Direct usage of a single existing state-of-the-art methods utilising different augmentation techniques (e.g., \textit{DuyHUYNH}) or different backbones (e.g., \textit{mimykgcp}, \textit{qzheng5}) resulted in improved results compared to the original baseline methods, however, much lower than the top performing methods (see Table~\ref{tab:ead2020_segmentation} and Table~\ref{tab:edd2020_segmentation_singleframe}). 

\textit{Hypothesis IV: The choice of combinatorial networks that well synthesizes width, depth and resolution to capture optimal receptive field, and a domain agnostic knowledge transfer mechanism are critical to tackle heterogeneous (multi-center and variable size) multi-class object segmentation task.}

From Figure~\ref{fig:ead2020_segmentation_teams_comparison} it can be observed that the top three performing teams of the EAD2020 segmentation task (\textit{arnavchavan04}, \textit{VinBDI}, \textit{mouradai$\_$ox}) has high DSC value (0.538, 0.548 and 0.492 respectively) compared to most methods for the specularity class instance. It is to be noted that the specularities are often confused with either artifact or bubbles which makes them hard to differentiate. For the instrument, saturation and bubbles class instances (see Figure~\ref{fig:EAD_segmentation_results} a.), most methods obtained high performance compared to other classes (e.g., the top three teams obtained 0.853, 0.844, 0.848  for the instrument; 0.722, 0.758, 0.703 for the saturation; and 0.738, 0.693, 0.693 for the bubbles class instance, respectively), artefact (DSC $< 0.520$) was among the worst class for most teams and for the baseline methods. This is mostly due to the variable size of artefacts; and the bubbles class instance is predominantly confused with either artefact or the specularity class (see Figure~\ref{fig:EAD_segmentation_results} b.). {Additionally, due to small sized and sparsely scattered specularity or bubble regions in some cases (for e.g., 4th image from left in Fig.~\ref{fig:dataset_samples} (a)), the annotator variability for these samples can have affected method performances for these classes. While checking for such biases is beyond the conducted study, we refer to the work by~\cite{Rolnick:arXiv2017}. The authors suggested that in general deep learning models are capable of generalizing from training data where the correct labels are outnumbered by the incorrect ones. However, the authors also acknowledged that a decrease in performance is inevitable and necessary steps such as using larger batch size and downscaling learning rate can help mitigate these issues.} 

% TODO EDD2020: reference for noisy labels Rolnick:arXiv2017
Unlike the EAD2020, the EDD2020 segmentation task comprised of larger shaped regions and only a few classes confused (see \ref{fig:supplementaryaboutDatafigure2} b.). Most methods scored comparably high DSC values with over 75\% for most of the disease classes except for suspicious class by most of the team. However, Figure~\ref{fig:EDD_segmentation_results} (b) (top) shows that while majority of teams were able to segment NDBE class area, the teams either missed the HGD area or miss classified HGD as suspicious class instance. It is to be noted that there is a very subtle difference between the HGD and the suspicious region even for the expert endoscopists. Similar observation can be found for the segmentation of protruded structures (Figure~\ref{fig:EDD_segmentation_results} (b), bottom) where most methods confused the class with the polyp class and the top two teams (\textit{adrian}, \textit{sahadate}) classified it as cancer class. Looking up into our expert consensus notes we found that these samples had hard to reach agreement cases  (i.e., suspicious and HGD classes; and cancer and polyp region).  

\textit{Hypothesis V: Instead of hard scoring of predicted mask classes that penalizes the method performance heavily in presence of marginal visual difference between classes and variability due to existing expert consensus in the dataset, probability maps can be used to mitigate such problem. {Additionally, teams should be encouraged to report results for different batch size and learning rates for obtaining better insight regarding performance especially when datasets are prone to have some incorrect labels.}}

\section{Conclusion}
We provided a comprehensive analysis of the deep learning methods built to tackle two distinct challenges in the gastrointestinal endoscopy: a) artefact detection and segmentation and b) disease detection and segmentation. It has been possible by the crowd-sourcing initiative of the EndoCV2020 challenges. We have provided the summary of the methods developed by the top 17 participating teams and compared their methods with the state-of-the-art detection and segmentation methods. Additionally, we dissected different paradigms used by the teams and presented a detailed analysis and discussion of the outcomes. We also suggested pathways to improve the methods for building reliable and clinically transferable methods. {In future, we aim towards more holistic comparison of the built techniques for clinical deployability by testing for hardware and software reliability in clinical settings.} 
\section*{Acknowledgments}
The research was supported by the National Institute for Health Research (NIHR) Oxford Biomedical Research Centre (BRC). The views expressed are those of the authors and not necessarily those of the NHS, the NIHR or the Department of Health. S. Ali, B. Braden, A. Bailey and JE. East is supported by NIHR BRC and J. Rittscher by Ludwig Institute for Cancer Research and EPSRC Seebibyte Programme Grant (EP/M013774/1).~We want to acknowledge Karl Storz for co-sponsoring the challenge workshop. We would also like to acknowledge the annotators and our EndoCV2020 challenge workshop proceedings reviewers and IEEE International Symposium on Biomedical Imaging 2020 challenge committee Michal Kozubek and Hans Johnson.
\section*{Author contributions}
S. Ali conceptualized the work, led the challenge and workshop, prepared the dataset, software and performed all analyses. M. Dmitrieva, N. Ghatwary, and S. Bano served as organising committee and participated in annotations. A. Bailey, B. Braden, J.E. East, R. Cannizzaro, D. Lamarque, S. Realdon were involved in the validation and quality checks of the annotations used in this challenge. G. Plolat, A. Temizel, A. Krenzer, A. Hekalo, YB. Guo, B. Matuszewski, M. Gridach, V. Yoganand assisted in compiling the related work and method section of the manuscript. S. Ali wrote most of the manuscript with inputs from M. Dmitrieva, N. Ghatwary, S. Bano and all co-authors.  All authors participated in the revision of this manuscript and provided substantial input.
%\section*{Additional information}
%\textbf{Competing interests}
%The author(s) declare no competing interests.

%\section*{Acknowledgments}
%Acknowledgments should be inserted at the end of the paper, before the
%references, not as a footnote to the title. Use the unnumbered
%Acknowledgements Head style for the Acknowledgments heading.
%
%\section*{References}
%
%Please ensure that every reference cited in the text is also present in
%the reference list (and vice versa).

%\section*{\itshape Reference style}
%
%Text: All citations in the text should refer to:
%\begin{enumerate}
%\item Single author: the author's name (without initials, unless there
%is ambiguity) and the year of publication;
%\item Two authors: both authors' names and the year of publication;
%\item Three or more authors: first author's name followed by `et al.'
%and the year of publication.
%\end{enumerate}
%Citations may be made directly (or parenthetically). Groups of
%references should be listed first alphabetically, then chronologically.

%%Harvard
%\bibliographystyle{model2-names.bst}\biboptions{authoryear}
\bibliography{endocv2020}

\end{document}